# International Multi-Centre Validation of Unsupervised Domain Adaptation for Precise Discrimination between Normal and Abnormal Retinal Development

## Author Information

Zhanhan Tu MBChB, MM, PhD,[1] Nikhil Reddy MSc,[2] Helen J Kuht PhD,[1] Ha-Jun Yoon PhD,[3] Gail DE Maconachie PhD ,[4] Ravi Purohit BMedSci, PhD,[1] Prateek Pani MSc,[2] Line Kessel, MD, PhD,[5,6]Jinu Han MD,[7] Jing Jin MD,[8] Shui-Hua Wang PhD,[9] Karthik Bharadwaj MSc,[11] Viral Sheth PhD,[1] Michael Hisaund BMedSci,[1] Rebecca J McLean MSc, PhD,[1] George Garratt MBChB,[1] Chetan Kantibhai Patel FRCOphth,[12] Kanmin Xue FRCOphth, PhD,[13] Callum Hunt MSc,[1] Zisen Zhou PhD,[1] Ian De Silva FRCOphth,[14] Seema Teli BMedSci,[1] Sonal Shah BSc, MCOptom,[1] Vanessa Rodwell MBChB, MSc,[1] Aarti Patel MBChB, FRCOphth,[1] Irene Gottlob MD, FRCOphth,[1] Yu-Dong Zhang PhD,[9,10] Zedong Zheng PhD,[9] Girish Varma PhD,[2*] Mervyn G Thomas PhD, FRCOphth[1*]

## Affiliations

[1]The University of Leicester Ulverscroft Eye Unit, School of Psychology and Vision Sciences, University of Leicester, RKCSB, PO Box 65, Leicester, LE2 7LX, UK

[2]International Institute of Information Technology, Gachibowli, Hyderabad - 500 032, India

[3] inAmind Laboratory, Department of Neuroscience, Psychology and Behaviour, University of Leicester, George Davies Centre, Lancaster Road, Leicester, LE1 7RH, UK.

[4] Academic Unit of Ophthalmology and Orthoptics, University of Sheffield, Sheffield, UK

[5] Department of Ophthalmology, Copenhagen University Hospital, Rigshospitalet-Glostrup, Copenhagen, Denmark

[6]The Department of Clinical Medicine, University of Copenhagen, Denmark

[7] Institute of Vision Research, Department of Ophthalmology, Gangnam Severance Hospital, Yonsei University College of Medicine, Seoul, Korea; Foveal Development Investigators Group.

[8]Sidney Kimmel Medical College of Thomas Jefferson University, Philadelphia, Pennsylvania; Nemours Children's Health, Wilmington, Delaware.

[9]School of Computing and Mathematical Sciences, University of Leicester, LE1 7RH, UK.

[10] School of Computer Science and Engineering, Southeast University, Nanjing, Jiangsu 211102, China

[11]Université Grenoble Alpes, CNRS, Grenoble INP, Institut Néel, 38000, Grenoble, France.

[12] Oxford Eye Hospital, Oxford University Hospitals NHS Foundation Trust, Oxford, UK; Ophthalmology Department, Great Ormond Street Hospital for Children NHS Foundation Trust, London, UK.

[13]Nuffield Department of Clinical Neurosciences, University of Oxford, Oxford, UK; Oxford University Hospitals NHS Foundation Trust, Oxford, UK; Great Ormond Street Hospital for Children NHS Foundation Trust, London, UK.

[14] Department of Ophthalmology, University Hospitals of Leicester, Infirmary Square, LE1 5WW, Leicester, UK.

***Corresponding authors:** Tel: +44 (0)116 252 5879, Fax: +44 (0)116 223 1996, email:

gv53@le.ac.uk and mt350@le.ac.uk

## Abstract

**Background:** Arrested retinal development during childhood is a major cause of childhood visual impairment and blindness. Diagnosis is achieved by paediatric ophthalmologists using optical coherence tomography (OCT) together with standarised grading of foveal development. OCT based grading informs diagnosis, targeted genetic testing and also prognostication. We aimed to develop an artificial intelligence (AI) based system to accurately grade foveal development, addressing the challenge of variability across different studies, centres, and the diverse OCT devices and acquisition protocols they employ. Specifically, we used Unsupervised Domain Adaptation (UDA), to create a device and protocol-agnostic classification model distinguishing normal and abnormal retinal development. To enable discovery science approaches, we also tested this within the UK Biobank dataset.

**Methods:** This multicentre, retrospective cohort study used retinal OCTs across five centres in four countries (UK, Denmark, USA and South Korea) and the UK Biobank (UKB). This included diverse data obtained from four different OCT machines representing both table-mounted and handheld devices. The dataset was split into training (80%) and testing (20%) sets. We used UDA to generalise a classification model differentiating normal and abnormal retinal development in a device- and protocol-agnostic manner training a model on labelled images from a source device and unlabelled images from another target device, ensuring that the model adapts the target distribution to closely match the source distribution.

Foveal scans were acquired from four different manufacturers: (I) TM-OCT1, a table-mounted OCT (3037 scans); (II) HH-OCT1, a handheld OCT (2041 scans); (III) TM-OCT2, (2519 scans); and (IV) TM-OCT3 (242 scans). The foveal B-scans for (I) and (II) were extracted, annotated, and segmented from a dataset of >20,000 OCT scans, whereas TM-OCT2 scans (UK Biobank) were extracted from a database consisting of >85,771 volumetric images. Convolutional neural networks (CNN) backbone ResNet50 was assessed with each possible pair used as source (for supervised training) and target (for unsupervised training). The model's diagnostic accuracy was compared against clinician graders (7 graders with a range of experience: 1 to 10 years) using performance metrics including (i) sensitivity, (ii) specificity, and (iii) accuracy.

**Findings:** The cross-domain binary classification model achieved diagnostic accuracy of 87.75–96.06%, sensitivity of 88.68–98.38%, and specificity of 78.00–98.02% across validation sets. The Domain Adversarial Neural Network (DANN), trained on TM-OCT1 (labelled) and HH-OCT1 (unlabelled) images, demonstrated generalisability on the HH-OCT1 test set, with 96.06% accuracy, 98.38% sensitivity, and 89.11% specificity. The model achieved an Area Under Curve (AUC) of 0.994 (95% CI: 0.981-0.999), surpassing clinician-annotated benchmarks and exceeding state-of-the-art discriminative performance in cross-domain settings. Particularly, the DANN maintained robust accuracy despite domain shifts between imaging systems (TM-OCT1 to HH-OCT1), demonstrating adaptability to heterogeneous clinical environments. Its high sensitivity across stages suggests clinical utility in early detection, where conventional methods underperform.

**Interpretation:** We have demonstrated the use of a device-agnostic AI-based system in adult and paediatric OCT interpretation for developmental retinal disorders without extra-labelled data. The diagnostic performance of the AI-based system is comparable to that of a senior clinician (~10 years of experience). The proposed system has the potential to reduce inter-examiner variability and support clinical decision-making. Classifying normal and abnormal retinal development from OCT scans has both diagnostic and prognostic implications. We also apply this system within the UK Biobank dataset to demonstrate feasibility of deploying UDA at a large scale. This has potential for obtaining novel imaging-derived phenotypes from existing datasets for performing association studies and supporting discovery science approaches across domains. Our AI-based system provides the foundation for a real-time, point-of-use, diagnostic tool for retinal developmental disorders.

**Funding:** This study was supported by the Medical Research Council (MRC), London, UK (grant numbers MR/J004189/1, MRC/N004566/1 and MC_PC_17171); the Ulverscroft Foundation (supporting MGT and ZT); Fight for Sight (grant numbers 5009/5010 and 24NN181); the Medical Research Foundation and Wellcome Trust (grant numbers MRF-JF-EH-23-136 and 222530/Z/21/Z; supporting ZT); the National Institutes of Health (grant number DRF-2014-07-136; supporting AP); and the National Institute for Health and Care Research (grant number CL-2017-11-003; supporting MGT).

## Research in Context

### Evidence Before This Study

We searched PubMed, Scopus, Science Direct, and IEEE Explore from database inception to February 18th, 2024, for studies in English that applied machine and deep learning on optical coherence tomography (OCT) imaging for the identification of abnormal retinal development. We combined the search terms artificial intelligence ("AI" or "deep learning" or "machine learning"), OCT ("OCT" or "optical coherence tomography"), grading ("grading" or "classification" or "classify"), and foveal hypoplasia ("foveal hypoplasia" or "abnormal foveal development"). Studies in languages other than English that included an English abstract were also considered.

We found 1 study using machine learning from OCT imaging in foveal hypoplasia for visual acuity prediction. In our study, we set out to develop and validate a deep-learning-based approach for the automatic validation of foveal hypoplasia and its grading from OCT imaging. Prior to this study, the interpretation of OCT scans for grading arrested retinal development relied on manual assessment by clinicians, introducing subjectivity and variability. Moreover, using it across different datasets including large biobanks has not been previously performed. Existing evidence highlighted the need for a standardised, objective, and efficient method to assess retinal conditions in children. Previous attempts at using artificial intelligence (AI) in OCT interpretation often required extensive labelled datasets, limiting their applicability and scalability.

## Added Value of This Study

We developed a device-agnostic AI system using unsupervised domain adaptation (UDA) to grade arrested retinal development across heterogeneous OCT datasets, including the UK Biobank, without requiring device-specific labelled data. Our model achieved diagnostic accuracy comparable to senior clinicians (approximately 10 years experience) and demonstrated generalisability across OCT platforms (e.g., TM-OCT1, HH-OCT1). This system addresses domain shift, a key barrier to real-world AI deployment. It enables scalable, standardised OCT interpretation across diverse clinical and research settings.

## Implications of All Available Evidence

The proposed AI system achieves diagnostic accuracy comparable to experienced clinicians, performs consistently across different OCT devices, and gives the same output for the same scan, thereby avoiding the inter-examiner variability inherent in human interpretation. Together, these findings show that the system is accurate, generalisable and reproducible, which makes it a promising frontline screening tool for retinal developmental disorders. It can also be applied to large biobank datasets to support genome-wide and phenome-wide association studies. These results advance device-agnostic OCT interpretation and provide a foundation for integrating AI into routine clinical assessment. The next step is to establish how well these findings generalise across clinical practice, including different healthcare settings, patient populations and device types, and to evaluate the effectiveness of implementing the system within clinical care.

## Introduction

Arrested foveal development, or foveal hypoplasia, is a major cause of visual impairment and blindness in children. Retinal disease accounts for between 25% and 47% of severe paediatric visual impairment, and foveal hypoplasia is a common structural correlate across these conditions. Its importance has been underlined by international efforts such as the Foveal Development Investigators Group, which spans 21 centres across 12 countries [1]. Optical coherence tomography (OCT) is central to its assessment and allows the fovea to be imaged non-invasively, and the resulting scans inform diagnosis, direct targeted genetic testing and support prognostication [1]. Foveal development begins in mid-gestation and continues until around 13 years of age [2-4] . It proceeds through characteristic stages, including displacement of the inner retinal layers, migration of cone photoreceptors towards the developing fovea, and specialisation of the foveolar cones [5-7] . Disruption at any stage results in foveal hypoplasia, which is seen in albinism, prematurity, achromatopsia and in disorders arising from variants in PAX6, AHR and SLC38A8 [8-15] .

The Leicester grading system classifies foveal hypoplasia by the stage at which development is arrested [6] . The grade carries both diagnostic and prognostic value, and in preverbal infants it predicts future visual acuity [16]. Accurate grading depends on specialised expertise in paediatric OCT interpretation, which is scarce and unevenly distributed.

Deep learning offers a route to automating OCT interpretation and extending this expertise. Its use in this setting is limited by the diverse set of OCT instruments and protocols, which vary across centres and between table-mounted and handheld devices. This variation produces domain shift, in which the statistical distribution of images differs between the data a model is trained on and the data it later encounters. A model trained on one device therefore tends to perform poorly on another. Correcting this by retraining requires new labelled data for each device. For rare developmental disorders, where scans and expert annotation are both limited, this requirement is difficult to meet and constrains the use of automated grading in practice.

Automated grading that could be trusted across instruments, without relabelling for each new device,

would remove this barrier. Unsupervised domain adaptation (UDA) offers such an approach. It learns from labelled images on a source device and unlabelled images on a target device, and aligns their feature distributions so that a classifier transfers between instruments without any labelled data from the target. Here we develop and validate a device-agnostic model that classifies normal against abnormal retinal development from OCT. We test whether UDA maintains performance across different manufacturers, both table-mounted and handheld, and across cohorts differing in age and origin. We then compare the model against expert clinician graders. Such a system would provide a screening tool for retinal developmental disorders and a means of deriving standardised structural phenotypes at scale for genetic and epidemiological discovery.

# Methods

## Study design

This was a retrospective, multicentre study of retinal OCT images. Data were contributed by five centres in four countries, comprising University Hospitals of Leicester NHS Trust (UK), Oxford University Hospitals NHS Foundation Trust (UK), Nemours Children's Health (USA), Copenhagen University Hospital, Rigshospitalet (Denmark) and Yonsei University College of Medicine (South Korea), together with the UK Biobank. All contributing sites are part of the Foveal Development Investigators Group [1] . Images were acquired between January 2009 and January 2023 as part of routine care or research. The study addressed device-agnostic binary classification of normal against abnormal retinal development. Binary classification was chosen because its anticipated value lies in screening and in deriving imaging phenotypes at scale for discovery science within large datasets. Reporting followed the TRIPOD statement [17].

## OCT datasets and devices

Four datasets were used, spanning table-mounted and handheld instruments from four manufacturers (Table 1, Supplementary Table 1). TM-OCT1 was acquired on a Copernicus HR (table-mounted) at Leicester. HH-OCT1 was acquired on a Leica Envisu (handheld) at Leicester. TM-OCT2 was acquired on a Topcon 3D OCT 1000 Mk2 and drawn from the UK Biobank and Denmark. The fourth dataset was acquired on a Heidelberg Spectralis at three centres in the UK, USA and South Korea. TM-OCT1, HH-OCT1 and TM-OCT2 were used for model development. The Heidelberg dataset was reserved in its entirety as an external inference set and contributed to no stage of training or adaptation. Scan protocols for each device are given in Supplementary Table 1.

## Participants

We included patients with retinal developmental disorders and healthy controls with no evidence of abnormal retinal development. All diagnoses known to be associated with foveal hypoplasia were eligible. Given the rarity of these conditions, cases were not subcategorised by genotype, since retinal structure rather than genetic subgroup is the primary determinant of prognosis[3,6]. Longitudinal data were retained because retinal architecture changes over time in achromatopsia[9, 12, 18] and in other forms of foveal hypoplasia[19]. Scans were excluded where the fovea was not clearly visible due to poor quality. Participants below one year of age were excluded, because the outer retinal structures remain immature and foveal specialisation is incomplete at that age, precluding reliable grading [2, 4]. Demographic characteristics and case mix for each dataset are summarised in Table 1.

## Image preprocessing and data partitioning

For the three development datasets, images were split 80% to 20% into training and test sets at the level of the individual. All longitudinal timepoints and all parafoveal scans from a given participant were confined to a single split, so that no individual contributed to both training and test data. Parafoveal and perifoveal scans can resemble different grades of foveal hypoplasia, so for TM-OCT1 and HH-OCT1 we increased diversity by including parafoveal scans that mimic arrested foveal development, an approach we refer to as parafoveal augmentation. TM-OCT2 and the Heidelberg set did not include parafoveal augmentation. Images were preprocessed using standard methods. During training on TM-OCT1, HH-OCT1 and TM-OCT2 we applied random cropping to 224 by 224 pixels, flipping on the vertical axis, resizing to 256 by 256 pixels and normalisation using ImageNet mean and standard deviation. The Heidelberg set was used in full for testing and received no augmentation. An overview of the study design is shown in Figure 1.

**Fig. 1: The Overview of Study Design**

This figure summarises the design of a device-agnostic AI model trained to classify normal and abnormal retinal development using optical coherence tomography (OCT) scans. For example, labelled data from the TM-OCT1 system and unlabelled data from the HH-OCT1 system were integrated via unsupervised domain adaptation (UDA), enabling the model to harmonise differences between imaging devices. Performance was validated internally (TM-OCT1) and externally (HH-OCT1), with accuracy, sensitivity, and specificity compared to expert clinician assessments.

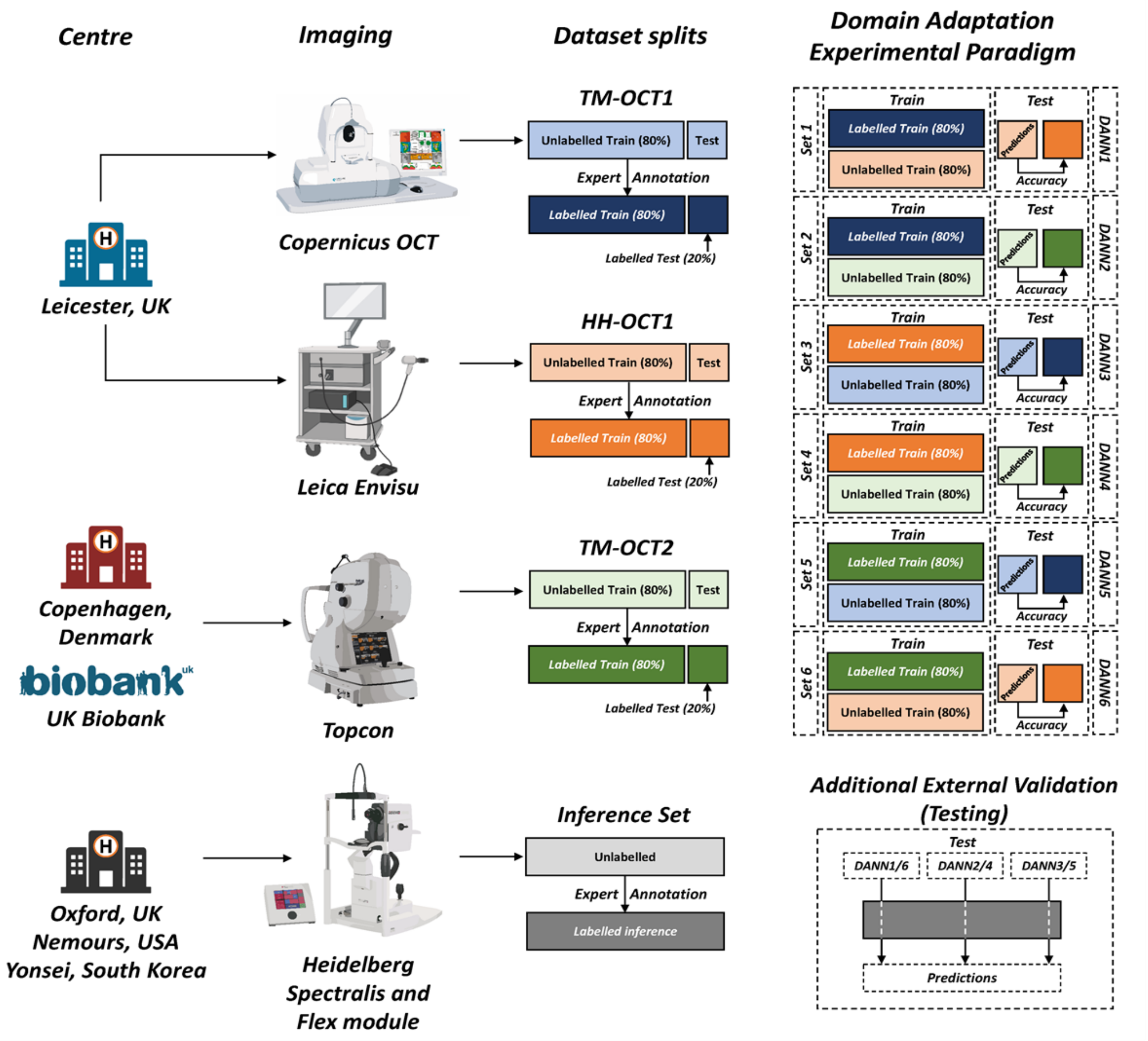

**Table 1 Demographic and clinical characteristics of participants stratified by diagnosis.**

Data are presented as median (IQR) or n (%).

| Characteristic | Control (n=477) | Albinism (n=108) | IIN (n=42) | INS (n=47) | Isolated FH (n=19) | Other (n=14) | Total (n=707) |
|---|---|---|---|---|---|---|---|
| **Age, years, median (IQR)** | 31.0 (20.0-45.0) | 24.0 (17.0-42.2) | 20.0 (11.2-42.0) | 20.0 (7.0-35.5) | 41.0 (29.5-48.0) | 34.5 (11.0-47.5) | 30.0 (19.0-44.0) |
| **Male, n (%)** | 211 (44.2) | 64 (59.3) | 22 (52.4) | 28 (59.6) | 10 (52.6) | 6 (42.9) | 341 (48.2) |
| **Female, n (%)** | 258 (54.1) | 44 (40.7) | 20 (47.6) | 19 (40.4) | 8 (42.1) | 8 (57.1) | 357 (50.5) |
| **Unknown sex, n (%)** | 8 (1.7) | 0 (0.0) | 0 (0.0) | 0 (0.0) | 1 (5.3) | 0 (0.0) | 9 (1.3) |
| **White, n (%)** | 361 (75.7) | 93 (86.1) | 38 (90.5) | 41 (87.2) | 17 (89.5) | 14 (100.0) | 565 (79.9) |
| **Asian or Asian British, n (%)** | 106 (22.2) | 11 (10.2) | 1 (2.4) | 4 (8.5) | 1 (5.3) | 0 (0.0) | 123 (17.4) |
| **Black, African, Caribbean or Black British, n (%)** | 2 (0.4) | 4 (3.7) | 0 (0.0) | 0 (0.0) | 0 (0.0) | 0 (0.0) | 5 (0.7) |
| **Other ethnicity, n (%)** | 3 (0.6) | 0 (0.0) | 0 (0.0) | 0 (0.0) | 0 (0.0) | 1 (7.1) | 4 (0.6) |
| **Undisclosed ethnicity, n (%)** | 5 (1.0) | 0 (0.0) | 3 (7.1) | 2 (4.3) | 1 (5.3) | 0 (0.0) | 10 (1.4) |

## Phenotype and grading

Foveal development was graded using the Leicester grading system for foveal hypoplasia (Figure 2) [6] . We used the clinical grade recorded at the point of care, with independent validation of each grade by expert investigators (HJK, ZT). Disagreements were resolved by a third senior grader. Where no grade was recorded in the case notes, a consensus grade was assigned by three investigators (GG, ZT, HJK). For binary classification, grades 1 to 4 and atypical foveal hypoplasia were combined into a single abnormal class, classified against normal foveal development.

**Fig. 2: Leicester grading system of foveal hypoplasia**

(A) and (C) Stages of foveal development and corresponding grades of foveal hypoplasia (B) A diagrammatic representation of the foveal hypoplasia grading scheme,4 with the corresponding real OCT examples representing the respective foveal hypoplasia grades. Abbreviations: RNFL = retinal nerve fibre layer; GCL = ganglion cell layer; IPL = inner plexiform layer; INL = inner nuclear layer; OPL = outer plexiform layer; ONL = outer nuclear layer; ELM; external limiting membrane; IS = inner segment; OS = outer segment; RPE = retinal pigment epithelium. Adapted and reproduced with permission from Thomas et al.

(2011)

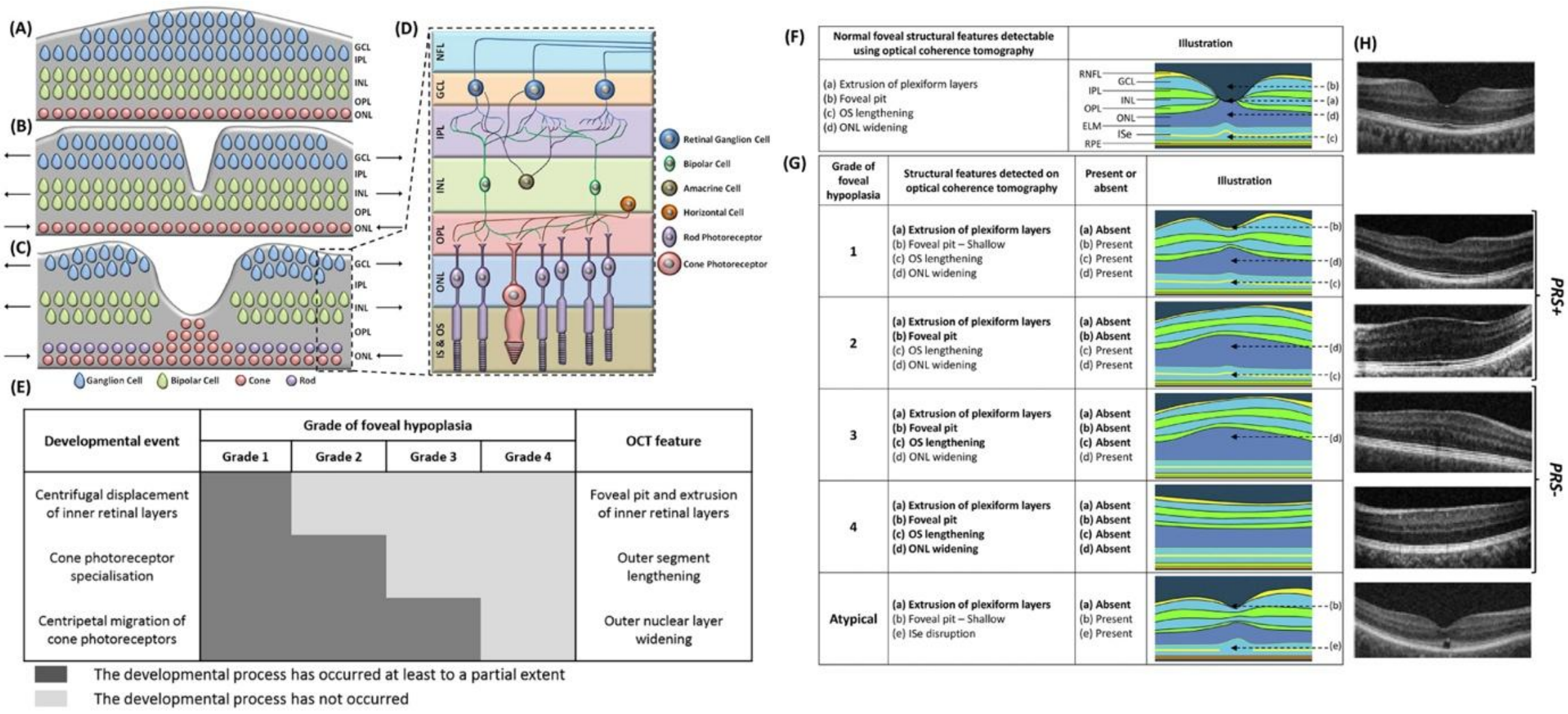

## Model development and domain adaptation

We first trained a baseline classifier without adaptation. This used a ResNet50 backbone trained on a labelled source device and applied directly to an unlabelled target device, to quantify the effect of device shift.

We then applied unsupervised domain adaptation using Domain Adversarial Neural Networks (DANN) [20]. A DANN learns from labelled source images and unlabelled target images, and aligns their feature distributions so that the learned features are informative for classification but uninformative about the device of origin. We developed models pairwise across TM-OCT1, HH-OCT1 and TM-OCT2, using each dataset in turn as source and another as target, giving six models in total.

The architecture combined a ResNet50 feature extractor with a classifier and a domain discriminator (Figure 3). Features were extracted by ResNet50 followed by global average pooling, yielding a tensor of size 2048. The classifier was a single linear layer mapping 2048 features to one output. The domain discriminator was a three-layer network with shapes 2048 to 1024, 1024 to 1024 and 1024 to one. A gradient reversal layer connected the feature extractor to the discriminator. During backpropagation it reversed the sign of the domain gradient, which drove the feature extractor toward device-invariant representations. Training minimised a combined loss $L = Lc - \lambda Ld$, where $Lc$ is the classification loss computed on labelled source data, $Ld$ is the domain classification loss computed on source and target data, and $\lambda$ weights the two terms. Models were trained on the ALICE High Performance Computing facility. The three best-performing models were then applied to the Heidelberg inference set for external validation.

**Fig. 3: Domain Adversarial Neural Network (DANN) Training Architecture**

Domain Adversarial Neural Network (DANN) Training Architecture utilised for OCT scan classification. The feature extractor network (F) captures discriminative features from OCT images, which are fed into two distinct pathways: the classifier network (C) and the discriminator network (D). The classifier (C), shown in red, uses labelled source data to categorize scans as Abnormal or Normal based on the class loss (Ly). In parallel, the discriminator (D), indicated in yellow, determines the domain origin (source or target) of the scans using unlabelled target data, informed by the domain loss (Ld). The Gradient Reversal Layer (GRL) connects F and D, facilitating the minimisation of domain discrepancies by reversing the gradient sign during backpropagation. The red and blue arrows represent the supervised and unsupervised training phases, respectively.

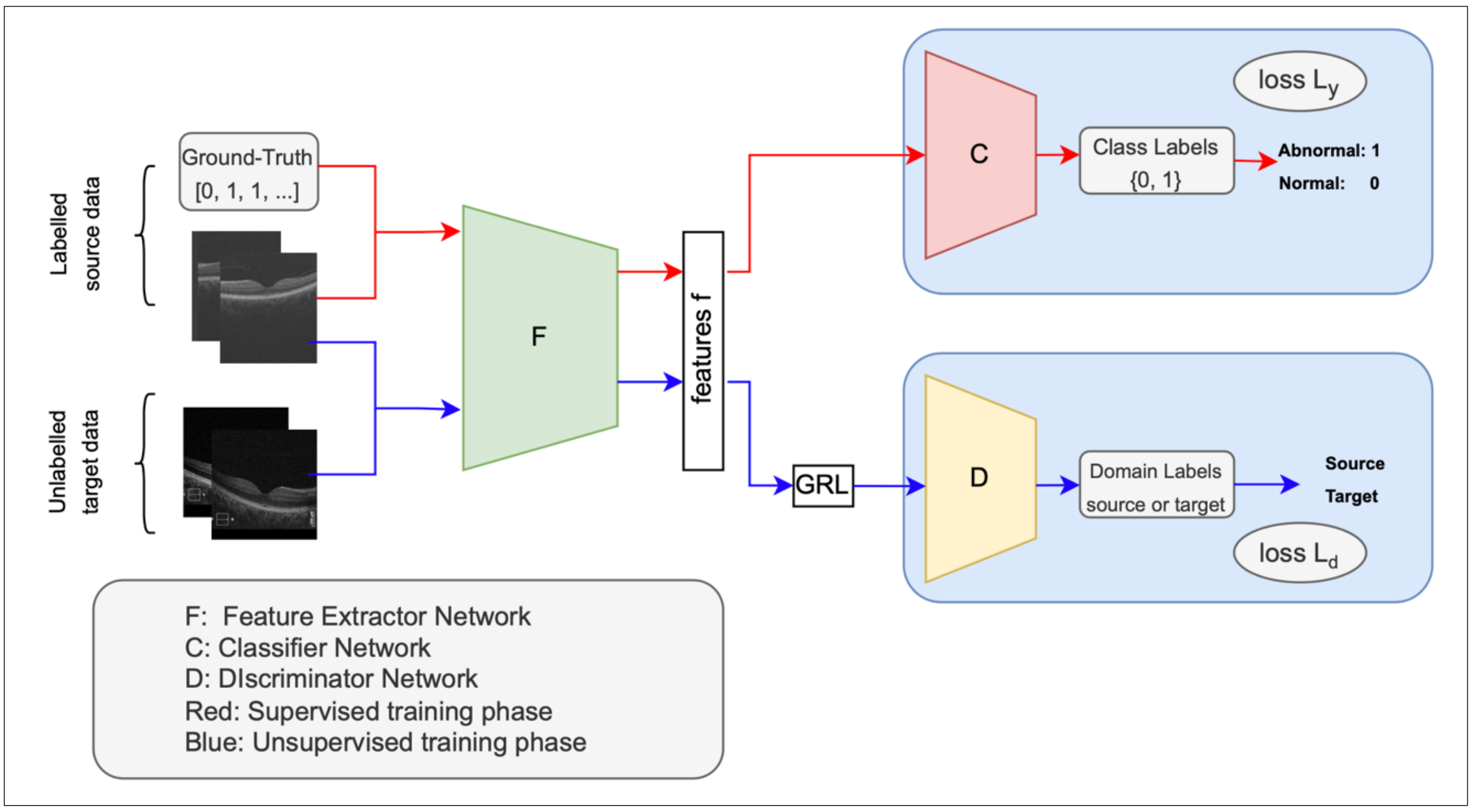

## Comparison with clinician graders

We compared model performance against 7 expert graders with 1 to 10 years of grading experience, using the same test images on which model accuracy was determined. Graders were blinded to all other clinical information and to one another's assessments. Each grader classified every scan as normal or abnormal.

## Statistical analysis

The primary endpoint was classification accuracy across datasets. We assessed each model using sensitivity, specificity, area under the receiver operating characteristic curve (AUC) and F1 score, and visualised performance using ROC curves. Confidence intervals were obtained by bootstrap resampling with 1,000 replicates for each source-to-target pairing, computing the false-positive and true-positive rates across thresholds to derive 95% confidence intervals. For the clinician comparison we summarised each grader's accuracy, sensitivity and specificity, quantified agreement among graders using Fleiss kappa overall and by experience level, and compared the best model against individual graders using McNemar's test on paired classifications. Tests were two-sided and considered significant at a threshold of 0.05. Analyses were performed in Python version 3.9.

# Results

## Cohort and study design

We assembled a retrospective, multicentre OCT dataset from five centres across four countries (UK, Denmark, USA and South Korea) and the UK Biobank, imaged between January 2009 and January 2023 on four instruments representing both table-mounted and handheld devices (Table 1). Development used three datasets, TM-OCT1 (Copernicus HR, table-mounted; Leicester, UK), HH-OCT1 (Leica Envisu, handheld; Leicester, UK) and TM-OCT2 (Topcon 3D OCT 1000 Mk2; UK Biobank and Denmark). A fourth dataset, acquired on a Heidelberg Spectralis at three centres (UK, USA and South Korea), was held out in its entirety as an external inference set. For the three development datasets, data were split 80% to 20% for training and testing at the level of the individual, with every longitudinal timepoint and every parafoveal B-scan from a participant confined to a single split.

The cohorts differed by design in age, device and case mix, providing a deliberately heterogeneous testbed for device-agnostic classification (Table 1). Foveal abnormality was graded following the Leicester system6, with grades 1 to 4 and atypical foveal hypoplasia combined into a single abnormal class and classified against normal foveal development.TM-OCT1 comprised 620 individuals (mean age 33.7 years, 63.0% abnormal) and HH-OCT1 a younger paediatric cohort of 174 individuals (mean age 12.2 years, 58.6% abnormal), both from Leicester and both including longitudinal data and parafoveal augmentation. TM-OCT2 drew on an older population (mean age 55.6 years, 71.4% abnormal), and the Heidelberg inference set comprised 242 scans from an ethnically diverse population (77.3% abnormal). Neither TM-OCT2 nor the Heidelberg set included longitudinal data or parafoveal augmentation.

## Unsupervised domain adaptation overcomes device shift

Models were built and compared across the three development datasets, TM-OCT1, HH-OCT1 and TM-OCT2. Each was used in turn as the labelled source and another as the target, giving six source-to-target pairings. We first established a baseline with a ResNet50 classifier trained on one device and applied directly to another, without adaptation. This baseline performed poorly. It over-predicted the abnormal class, with a false-positive rate of 0.95 and an F1 score of 0.86 on the target device. A classifier accurate on its own instrument therefore became unreliable once the acquisition hardware changed. To address this, we applied unsupervised domain adaptation (UDA). UDA aligns the feature distributions of a labelled source device and an unlabelled target device through a domain-adversarial objective. This allows a classifier to transfer between instruments without any labelled data from the target and without additional annotation.

Adaptation closed much of the gap. The false-positive rate fell from 0.95 to 0.11 and the F1 score rose from 0.86 to 0.97. The strongest pairing was a DANN trained on labelled TM-OCT1 and unlabelled HH-OCT1 images. On the held-out HH-OCT1 test set it reached 96.06% accuracy, 98.38% sensitivity and 89.11% specificity, having seen no HH-OCT1 label during training. Across all three pairings, accuracy ranged from 87.75% to 96.06%, sensitivity from 88.68% to 98.38% and specificity from 78.00% to 98.02% (Table 2, Supplementary Figure 1, Supplementary Table 2). Performance varied with the source device. Table-mounted TM-OCT1 generalised well, reaching 96.06% on HH-OCT1 and 92.64% on TM-OCT2. The handheld HH-OCT1 as source was weaker, at 87.74% and 88.86%, consistent with noisier handheld images providing a less transferable source distribution. The choice of source also shaped the error profile. One model adapted to TM-OCT2 showed a lower false-positive rate and a higher false-negative rate than the other (Figure 4, other model performance in Supplementary Figure 2). This suggests the approach can be tuned toward sensitivity or specificity according to clinical need. Domain adaptation therefore allowed a model that had failed on unseen hardware to perform consistently across instruments without further labelling.

**Table 2: Classification metrics for Unsupervised Domain Adaptation on source-target pairs**

TPR: True Positive rate, FPR: False Positive Rate; Sensitivity = TP/(TP+FN); Specificity =; TN/(FP+TN); Precision = TP/(TP+FP); False Positive Rate = FP/(FP+TN); False Discovery Rate (FDR) = FP/(FP+TP); False Negative Rate (FNR) = FN/(FN+TP); F1score = 2*TP/(2*TP+FP+FN)

| DANN | SOURCE TO TARGET | Accuracy | Sensitivity TP/(TP+FN) | Specificity TN/(FP+TN) | Precision TP/(TP+FP) | False positive Rate (FPR) FP/(FP+TN) | False Discovery Rate (FDR) FP/(FP+TP) | False Negative Rate (FNR) FN/(FN+TP) | F1 score 2*TP/(2*TP+FP+FN) |
|---|---|---|---|---|---|---|---|---|---|
| DANN1 | TM-OCT1 – HH-OCT1 | 0.95 | 0.96 | 0.93 | 0.97 | 0.07 | 0.03 | 0.04 | 0.97 |
| DANN2 | TM-OCT1 – TM-OCT2 | 0.92 | 0.92 | 0.93 | 0.97 | 0.08 | 0.03 | 0.08 | 0.94 |
| DANN3 | HH-OCT1 – TM-OCT1 | 0.87 | 0.89 | 0.85 | 0.91 | 0.15 | 0.09 | 0.12 | 0.89 |
| DANN4 | HH-OCT1 – TM-OCT2 | 0.88 | 0.92 | 0.78 | 0.91 | 0.22 | 0.09 | 0.08 | 0.92 |
| DANN5 | TM-OCT2 – TM-OCT1 | 0.93 | 0.95 | 0.89 | 0.94 | 0.11 | 0.06 | 0.05 | 0.94 |
| DANN6 | TM-OCT2 – HH-OCT1 | 0.97 | 0.97 | 0.94 | 0.98 | 0.06 | 0.02 | 0.03 | 0.98 |

**Fig. 4: DANN2 vs DANN4**

The figure illustrates the performance of Domain-Adversarial Neural Network (DANN) models (DANN2 and DANN4) in reducing diagnostic errors. Both models were test data set collected from TM-OCT2. DANN2 demonstrates statistically significant lower false positive (FP) and higher false negative (FN) rates compared to DANN4, as indicated by the reduced height of corresponding bars. Error bars represent variability across multiple validation trials.

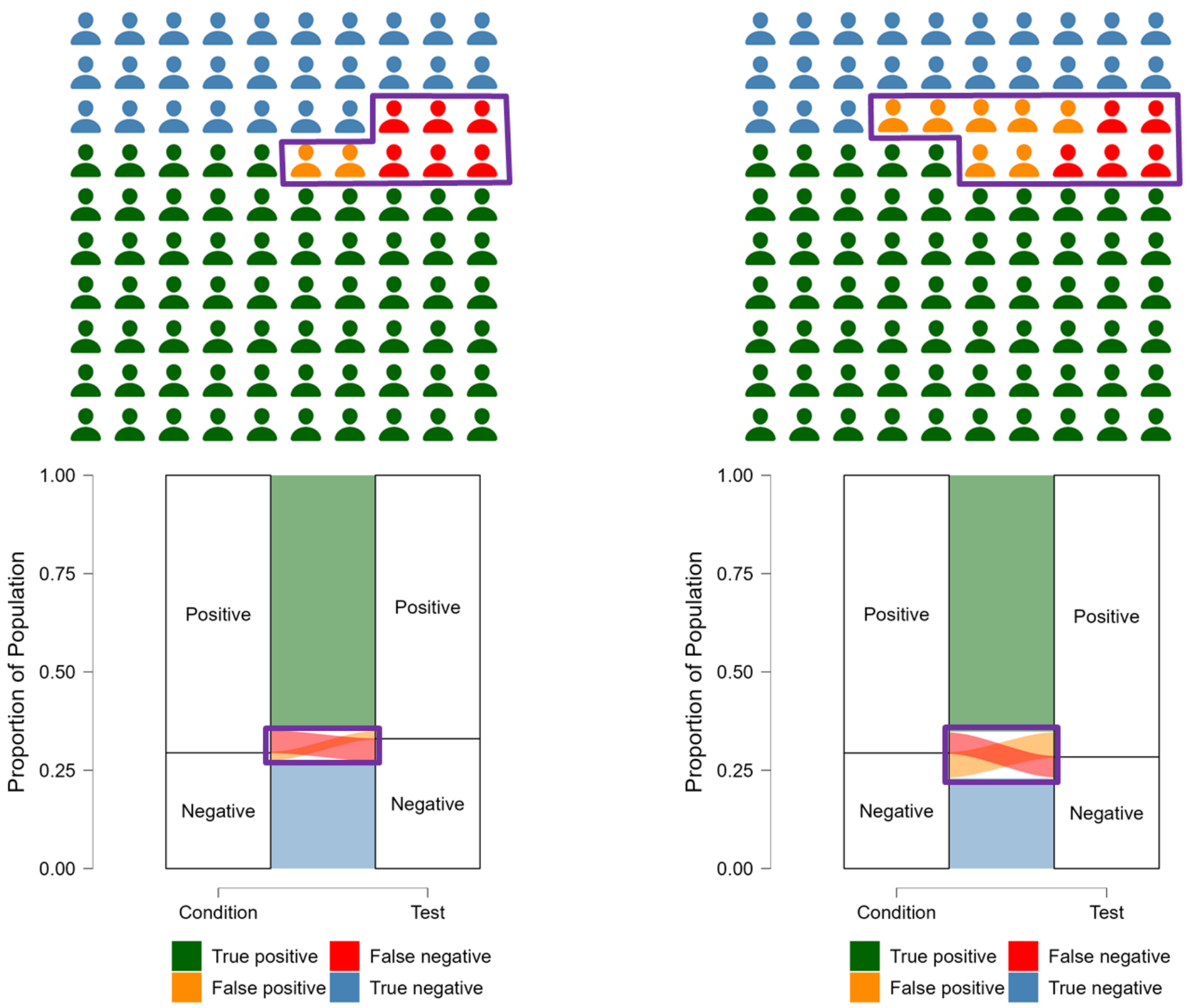

## Generalisation to an unseen device and new populations

Having shown that adaptation worked on each target device, we next asked whether it transferred to a device the models had never seen. For this we used a held-out inference set that played no part in training or adaptation. It comprised 242 Heidelberg Spectralis scans (187 abnormal, 55 normal) from a manufacturer not used in development, contributed by three centres in the UK, USA and South Korea. The best model for each target device (DANN1, DANN2 and DANN5) performed strongly on its own target test set, with F1 scores of 0.97, 0.94 and 0.94 and areas under the curve of 0.994 (95%

CI 0.981 to 0.999), 0.973 and 0.973 (Figure 5A). We then applied all six models to the Heidelberg set (Figure 5B, Supplementary Table 3). DANN1 to DANN4 maintained discrimination on this unseen device and in new patient populations, with AUCs between range 0.954 to 0.976. DANN5 and DANN6 performed worse, with AUCs of 0.871 and 0.793.

**Fig. 5: Comparative ROC Curves for Domain Adaptation between various Imaging Devices**
The ROC labelled (A1) TM-OCT1(source-domain) to HH-OCT1(target-domain) was the best performing model when adapting to HH-OCT1. Similarly, (A2) and (A3) were best models adapting to TM-OCT2 and TM-OCT1(as targets). Fig (B), plot shows ROC curve for the 6 models.

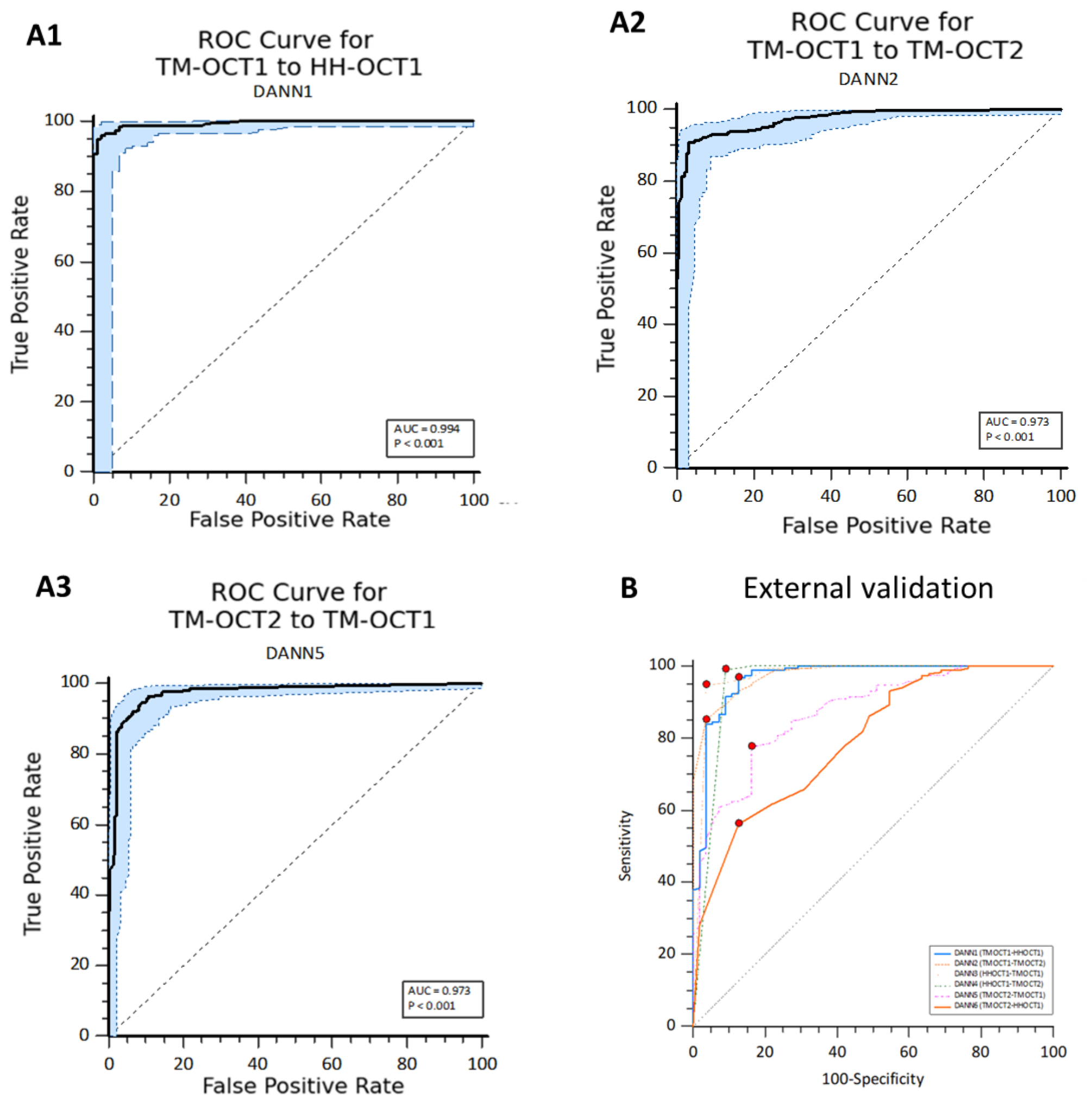

## Comparison with expert clinician graders

Having shown that the model generalised across devices, we next compared its performance with that of seven expert clinician graders to determine whether it reached the standard of routine clinical grading. Accuracy, sensitivity and specificity were calculated for each grader on the same test set, alongside their years of clinical experience. Inter-grader agreement was then assessed both overall and according to experience level to quantify variability between clinicians. The model was subsequently compared with individual graders and with the senior and junior grader groups, with paired statistical testing used to evaluate differences in performance. Overall, the model performed at the level of the most experienced graders and showed more consistent classification than was observed across clinicians with differing levels of experience (Figure 6, Supplementary Figure 3).

**Fig. 6: Performance of the AI-system demonstrated with receiver operating characteristic (ROC) curve for abnormal versus normal classification**

ROC of the best model showing the confidence interval, along with the annotations of clinicians overlayed on top of the confidence interval band.

The below figure shows a zoomed-in version of the relevant part. We can see how the FPR-TPR, according to the annotations of clinicians with the least experience are not within the confidence band, but the ones with higher experience are within the band except for clinician 1.

ROC: Receiver Operating Characteristics, FPR: False Positive rate, TPR: True Positive Rate

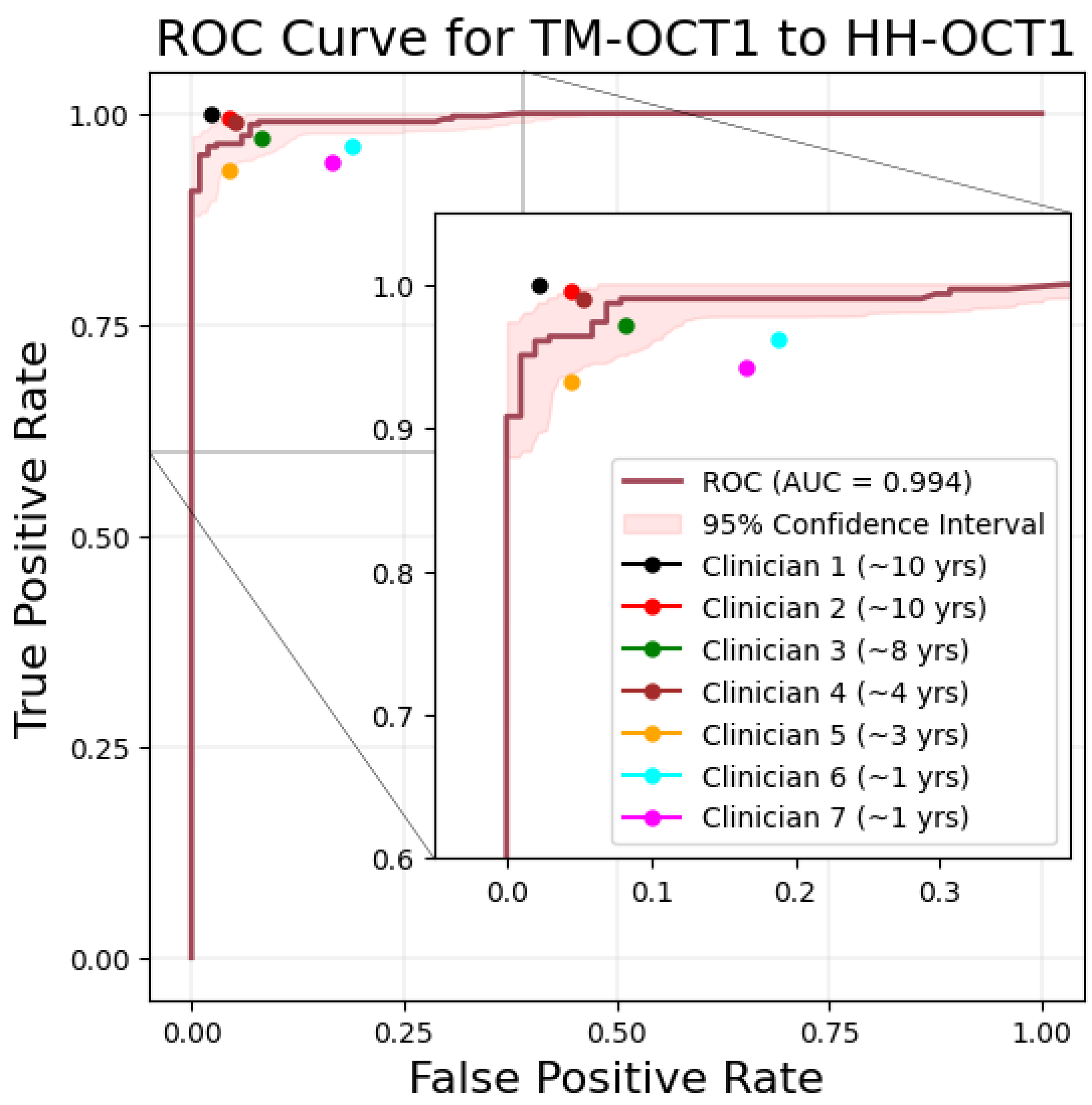

## Discussion

We developed a device-agnostic model that classifies normal against abnormal retinal development from OCT, using unsupervised domain adaptation to transfer between instruments without any labelled data from the target device. The model performed at the level of experienced clinician graders, and this performance held on an entirely unseen manufacturer and in new patient populations. A single adapted model therefore classified arrested retinal development reliably across table-mounted and handheld devices and across cohorts differing in age and origin.

A conventional classifier trained on one device failed once the acquisition hardware changed, and correcting this by retraining would require new labelled data for every new device. That requirement is difficult to meet for rare developmental disorders, where both scans and expert annotation are limited. By aligning the feature distributions of the source and target devices, UDA removed this dependence and allowed one model to generalise without further labelling. The result is a system that is accurate, generalisable and reproducible together, rather than accurate on a single instrument alone. This combination is what makes deployment across centres, and extraction of phenotypes from existing datasets, practical.

This property also clarifies what the model standardises. Clinicians apply consistent grading criteria, and the variability the model removes does not arise from disagreement over those criteria. It arises from differences in imaging hardware and acquisition protocols across centres, which alter the appearance of the same underlying anatomy. Our own work has shown that grading agreement varies with experience and imaging conditions [21], and it is this hardware-driven and experience-driven variability that a device-agnostic model absorbs. The model returns a standardised structural readout that is stable across the instruments on which foveal development is imaged.

Prior work in this specific area is limited. Machine learning applied to foveal hypoplasia from OCT has been reported only for visual acuity prediction, and a recent deep learning study graded foveal hypoplasia from colour fundus images with lower binary accuracy and no cross-device design [22]. Broader efforts in medical image domain adaptation have aligned feature distributions across scanners and protocols in other modalities [23-29], but none has addressed device-agnostic classification of retinal developmental disorders, nor validated transfer between table-mounted and handheld OCT. Our study extends domain adaptation to this setting and tests it on a held-out manufacturer, which is a stronger form of external validation than within-dataset testing.

A device-agnostic classifier could also serve as a screening tool where paediatric OCT expertise is scarce, flagging abnormal retinal development for expert review and supporting more equitable access to diagnosis. Because it transfers to handheld devices, it is applicable to the imaging of infants, in whom handheld OCT is often the only feasible option. Beyond individual diagnosis, the same model can generate standardised, device-independent structural phenotypes at scale. Applied across large imaging datasets, this offers a route to imaging-derived phenotypes for genome-wide and phenome-wide association studies, and provides a way to study retinal development in populations that are otherwise limited by inconsistent grading across instruments.

We acknowledge several limitations of this study. First, the model classifies individual 2D B-scans rather than volumetric OCT, so the spatial context of the full volume is not used, and extending domain adaptation to 3D volumes remains an open and non-trivial problem 22. Secondly, the model does not grade severity, and it does not distinguish a flat central macula caused by arrested foveal development from one caused by loss of the inner retina, for example following prenatal haemorrhage, nor does it identify the underlying gene or aetiology. These are boundaries of scope, but they define what the current system can and cannot support clinically. Thirdly, the reference standard was expert grading, so the model was validated against human labels rather than an independent biological measure. Fourthly, case mix and age differed substantially across cohorts, and the proportion of abnormal scans ranged from 58 to 77 percent, so accuracy should be interpreted alongside sensitivity and specificity, and residual confounding by age or sex cannot be excluded. Fifthly, the design was retrospective, ungradable scans were excluded, and there was no prospective or point-of-care evaluation, so performance on unselected, lower-quality scans acquired in routine practice remains to be established. Sixthly, handheld images were a weaker source domain, indicating that transfer is sensitive to the quality of the source data. Finally, the prognostic value of grading is bounded. The grade predicts future visual acuity in preverbal infants 16, but it does not indicate whether a condition will remain stationary or progress, as can occur with central atrophy in achromatopsia, so the prognostic use of a device-agnostic classifier should be framed as supporting acuity-related prognostication rather than predicting disease trajectory.

In summary, unsupervised domain adaptation enabled accurate, generalisable and reproducible classification of arrested retinal development across OCT devices, without any additional labelling. In doing so it offers both a practical screening tool for retinal developmental disorders and a foundation for standardised retinal phenotyping at scale, linking the clinic to population-level genetic and epidemiological discovery.

## Conclusion

Our study addresses a pivotal challenge in medical AI: enabling robust, device-agnostic diagnostic systems without the need for costly relabelling across heterogeneous imaging platforms. By pioneering the application of UDA to OCT, we developed an AI framework capable of accurately grading arrested retinal development across manufacturers, achieving diagnostic performance comparable to that of a senior clinician with over a decade of experience. Using a ResNet50 backbone, the model achieved cross-domain binary classification accuracies of 87.75–96.06%, with sensitivities of 88.68–98.38% and specificities of 78.00–98.02%. These results underscore UDA's potential to mitigate inter-device variability while maintaining clinical-grade reliability.

Significantly, our work represents the first preliminary validation for a domain-adaptive AI system in OCT interpretation, offering a scalable solution to avoid inter-examiner discrepancies and streamline clinical workflows. The ability of the framework to generalise across imaging devices without additional labelled data is particularly important as OCT becomes increasingly integrated into routine clinical practice and may facilitate the deployment of AI-assisted diagnosis across diverse imaging environments.

The implications extend beyond ophthalmology. By demonstrating capacity of UDA to reduce diagnostic errors (e.g., false positives) and enhance model generalisability, our findings advocate for broader adoption of domain adaptation techniques in medical imaging. This approach holds particular promise for resource-limited settings, where device heterogeneity and sparse expert annotations hinder equitable access to precision diagnostics. Future efforts should focus on validating these results in multi-centre trials and integrating 3D/4D imaging pipelines to further refine spatial-temporal feature extraction.

## Acknowledgements

We thank Basu Dawar, Swati Parida, Helena Lee, Teo Bica and Abdullah Aamir for their input with this study. Our sincerest thanks to John Kelly and Leica Microsystems who provided software for navigating and extracting OCT scans. This research used the ALICE High Performance Computing Facility at the University of Leicester. KX would like acknowledge funding from the Wellcome Trust (216593/Z/19/Z), NIHR Oxford Biomedical Research Council (NIHR203311) and Medical Research Council (MR/Z504725/1).

## Ethics approval

The study received ethical approval from the East Midlands, Leicester Central Research Ethics Committee (study title, Application of Artificial Intelligence in Ophthalmology; REC reference 20/EM/0040; protocol 0733) and Health Research Authority approval (IRAS ID 261121). UK Biobank recruitment and data acquisition were approved by the North-West Research Ethics Committee (reference 06/MRE08/75), and this project was approved by the UK Biobank (application ID 85881). The study followed the Declaration of Helsinki.

## Conflict of interest statement

All authors report no competing interests. The funders had no role in the design of the study; in the collection, analyses, or interpretation of data; in the writing of the manuscript, or in the decision to publish the results.

**Supplementary Table 1 OCT acquisition protocols**

Acronyms: Table-mounted OCT1 (TM-OCT1)= Copernicus OCT; Handheld OCT (HH-OCT1) = Leica Envisu; Table-mounted OCT2 (TM-OCT2) = Topcon OCT; Table-mounted OCT3 (TM-OCT3) = Heidelberg OCT.

| OCT | A-scans | B-scans | Scan Dimensions | Notes |
|---|---|---|---|---|
| TM-OCT1 | 743 | 75 | 7mm x 7mm | 3 trails per eye |
| | 500 | 100 | 10mm x 10mm | 2 protocols |
| | 500 | 100 | 10mm x 5mm | 2 protocols |
| HH-OCT1 | 600 | 80 | 12mm x 8mm | at least 2 trails per eye |
| TM-OCT2 | 512 | 128 | 6mm x 6mm (centre of fovea) | Topcon 3D OCT 1000 Mk2, same protocols |
| TM-OCT3 | N/A | N/A | 30-degree scan angle | High resolution mode, combination of multiple single and raster scans centred on the fovea (mixture) |

**Supplementary Figure 1 ROC Curves for Pairwise Comparisons of DANN Models**

ROC curves comparing pairs of models adapted to the same target device from different source devices. (A) Target HH-OCT1: DANN1 (source TM-OCT1) versus DANN6 (source TM-OCT2). (B) Target TM-OCT2: DANN2 (source TM-OCT1) versus DANN4 (source HH-OCT1). (C) Target TM-OCT1: DANN3 (source HH-OCT1) versus DANN5 (source TM-OCT2).

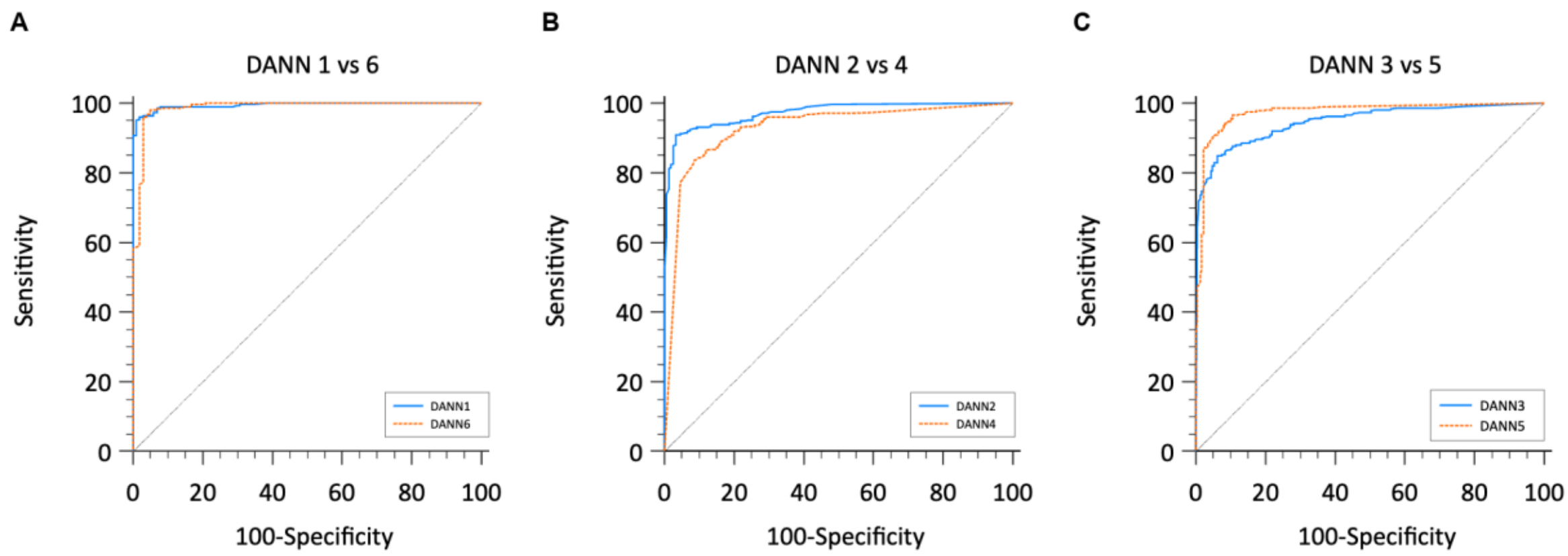

**Supplementary Table 2 Statistical Results of DANN Pairwise Comparisons**

| | P-value | Z-value | 95% Confidence Interval |
|---|---|---|---|
| **DANN 1 vs 6** | 0.2764 | 1.088 | -0.00589 to 0.0206 |
| **DANN 2 vs 4** | 0.0001 | 3.930 | 0.0227 to 0.0678 |
| **DANN 3 vs 5** | 0.0008 | 3.368 | 0.0108 to 0.0409 |

**Supplementary Figure 2 Classification outcomes of DANN1 to DANN6 on their target test sets**

## DANN1

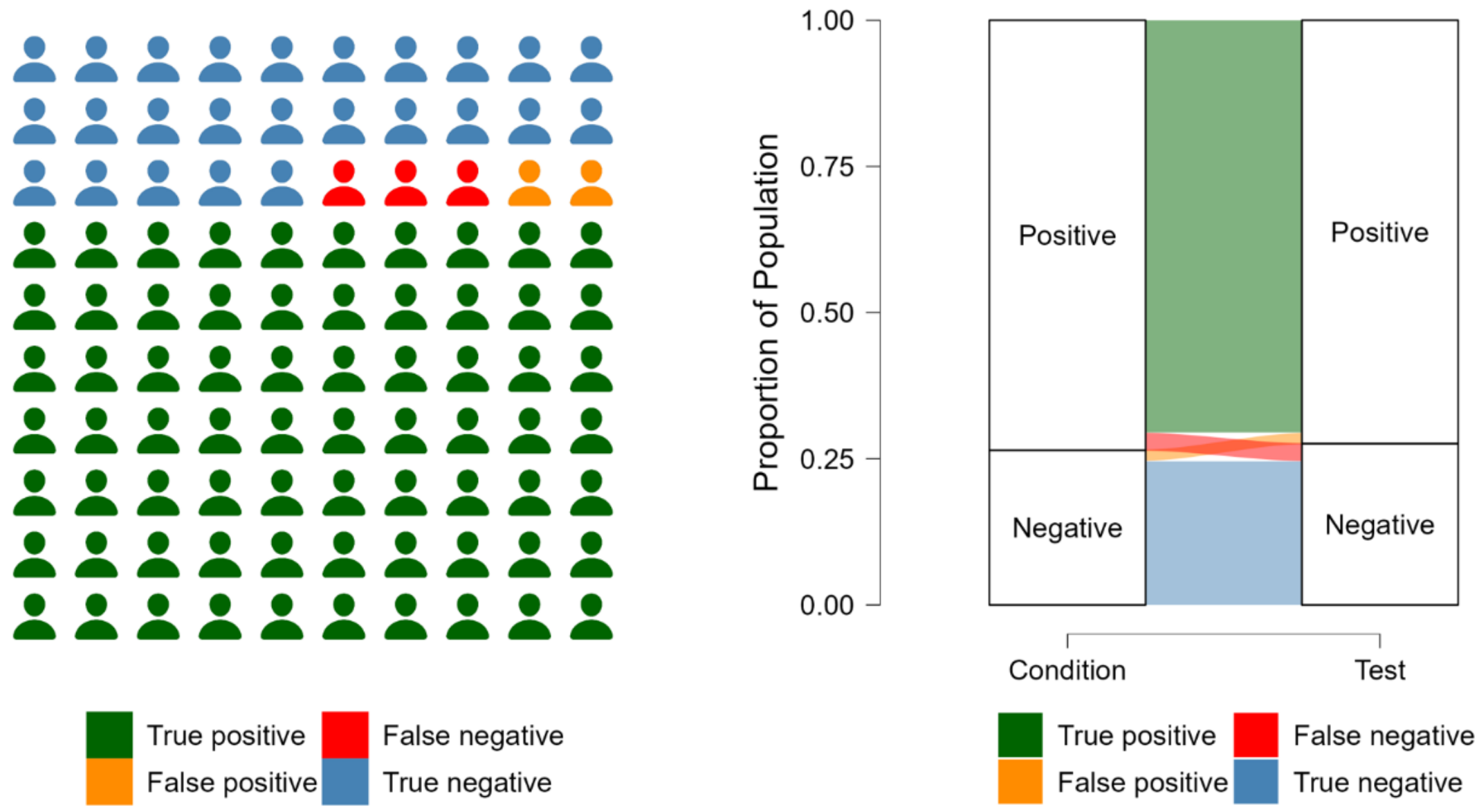

## DANN2

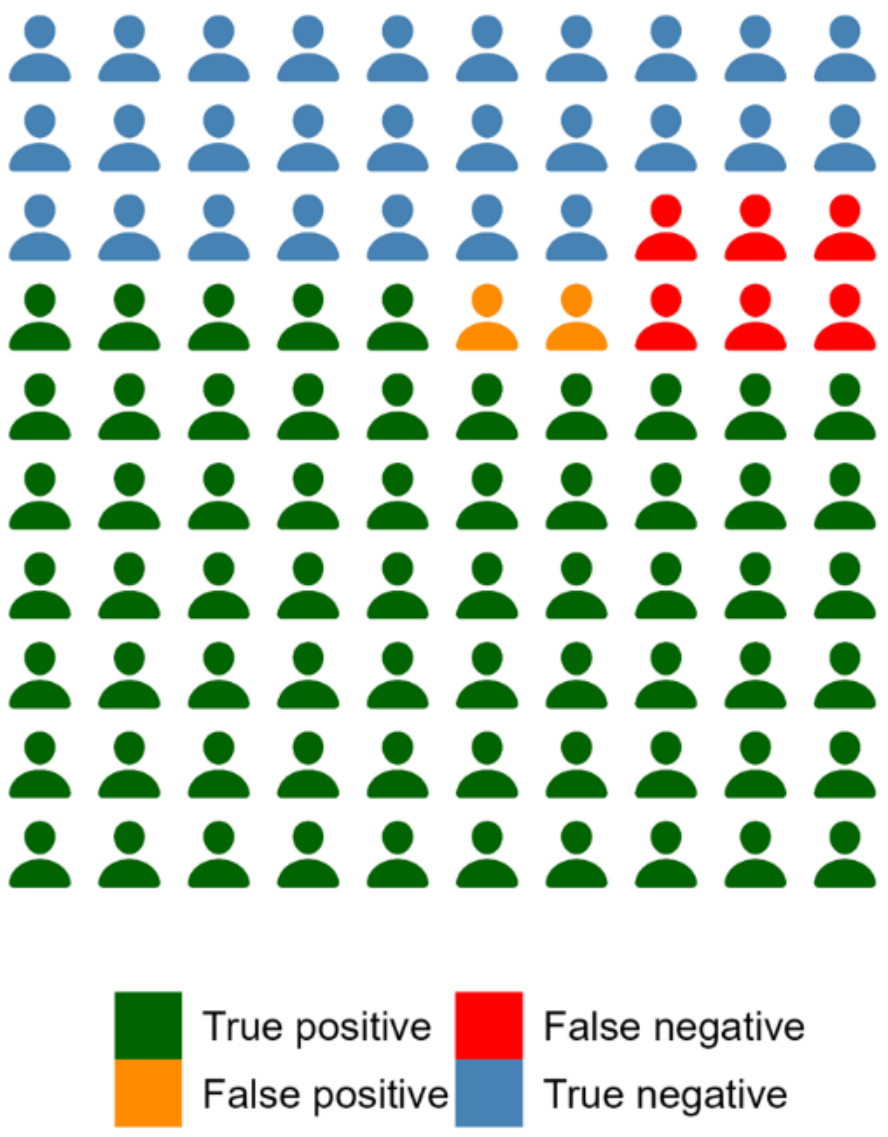

True positive
False negative
False positive
True negative

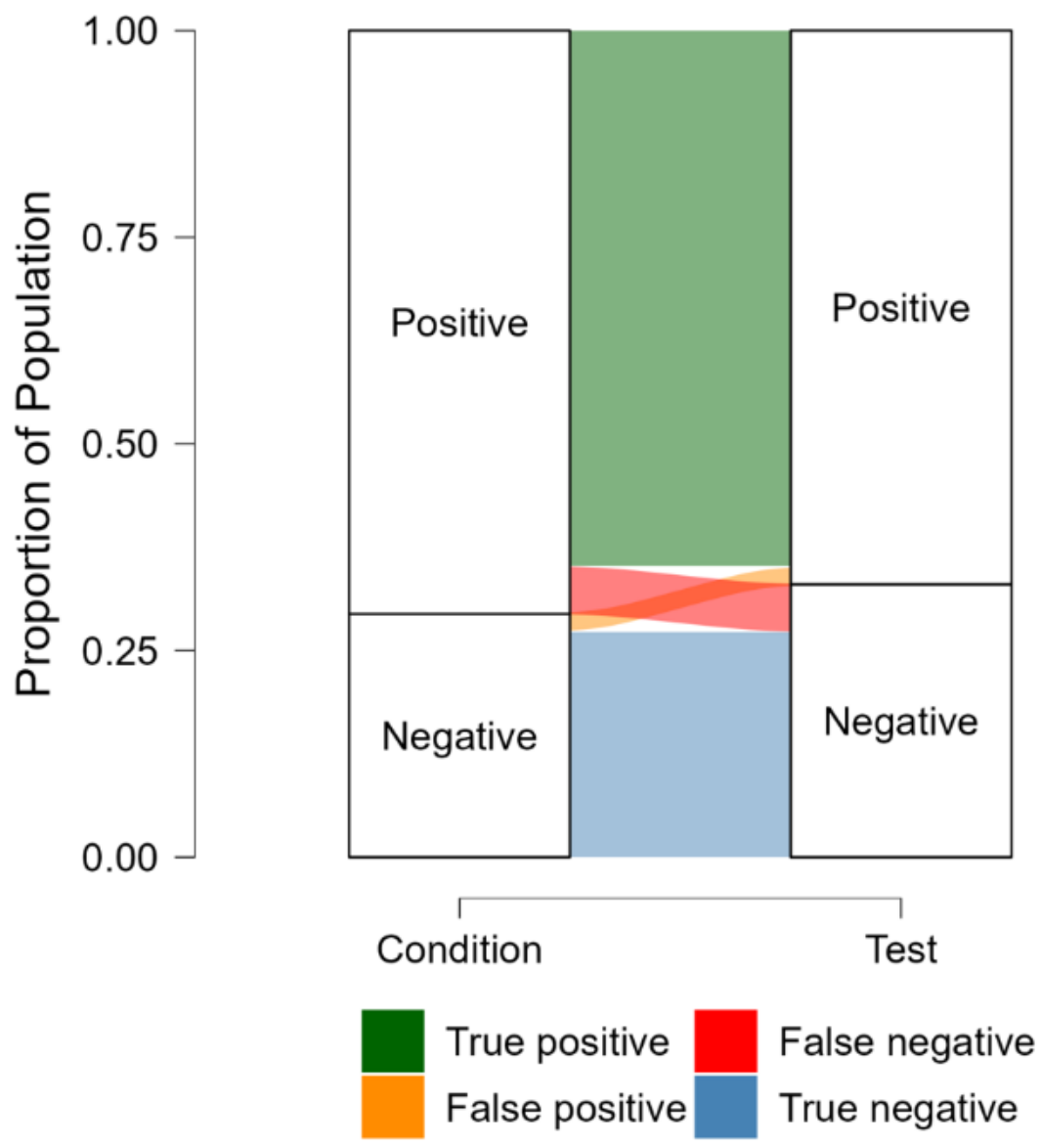

1.00
0.75
0.50
0.25
0.00
Proportion of Population
Positive
Negative
Positive
Negative
Condition
Test
True positive
False negative
False positive
True negative

## DANN3

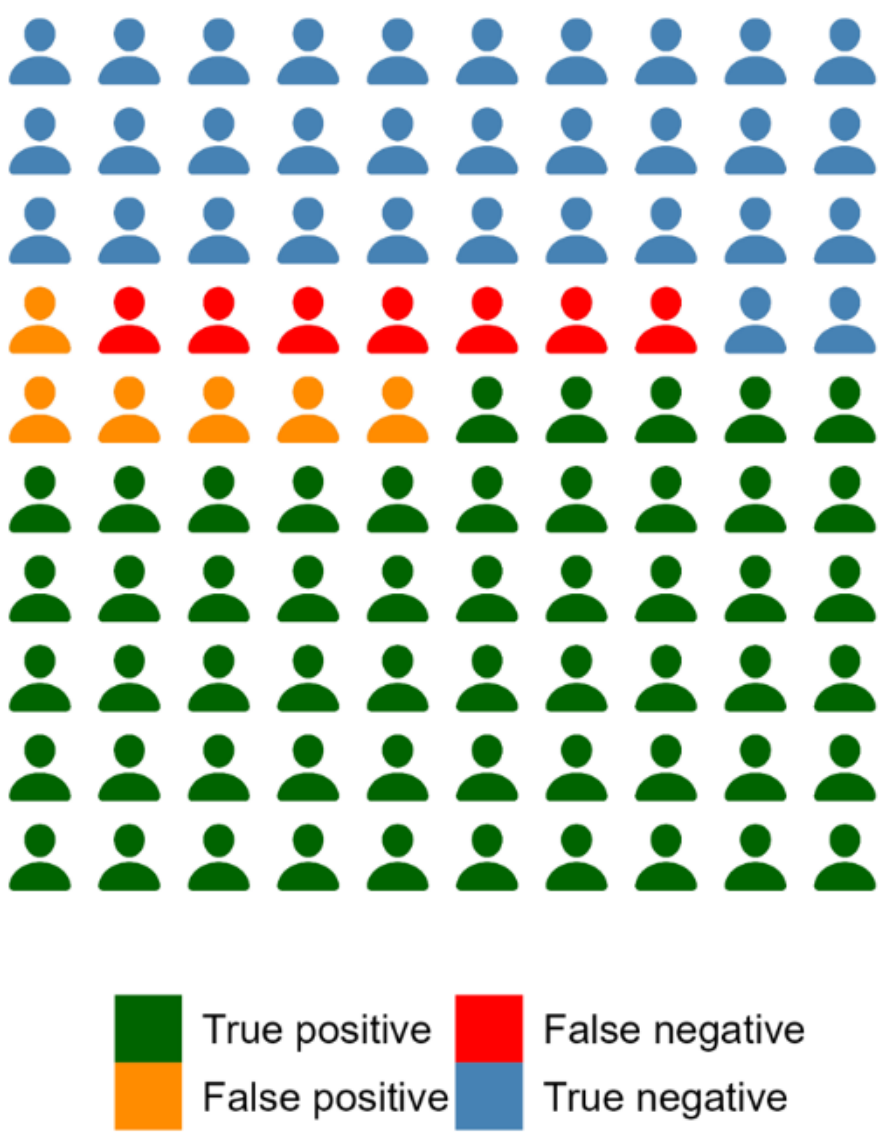

True positive
False negative
False positive
True negative

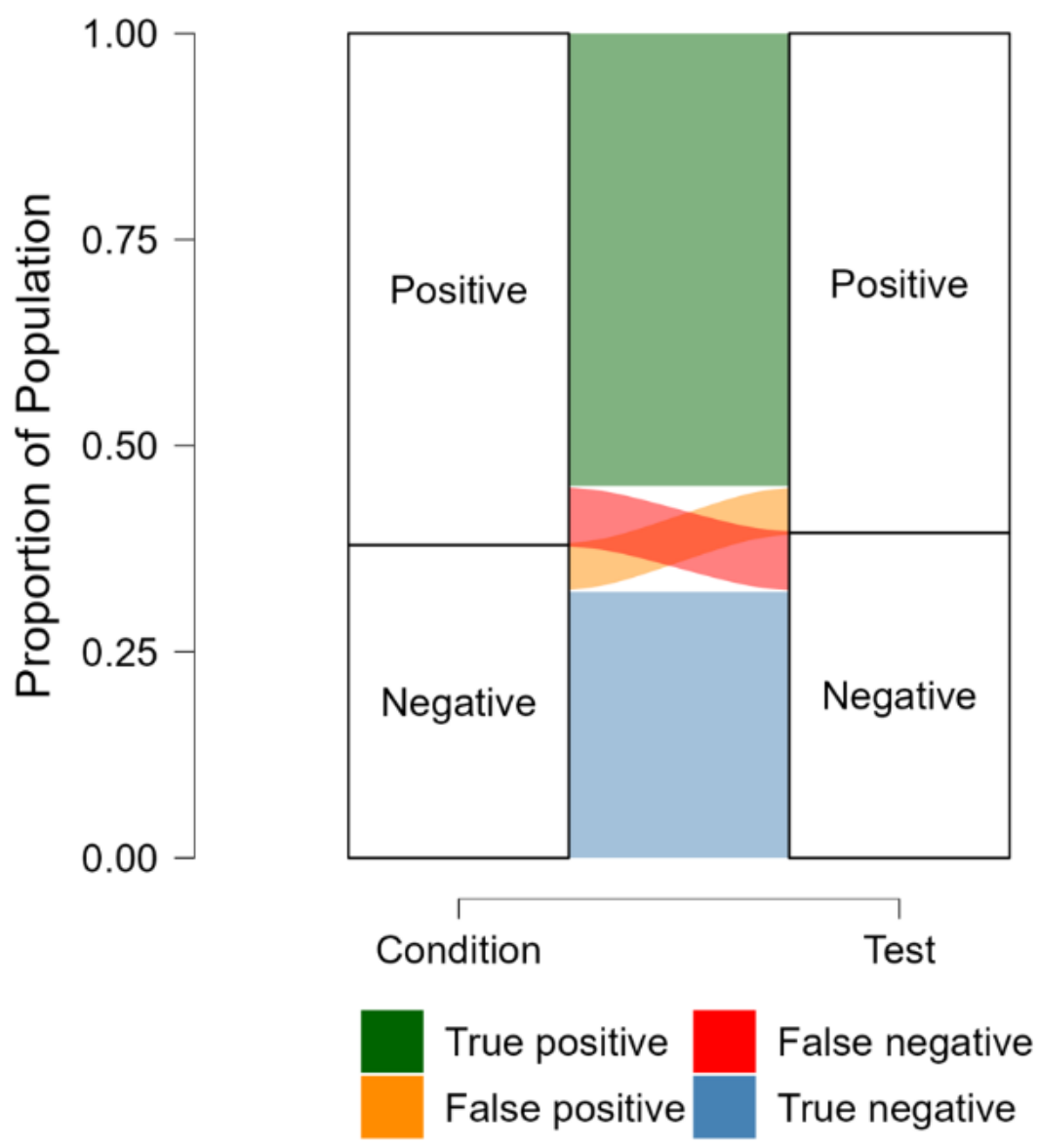

1.00
0.75
0.50
0.25
0.00
Proportion of Population
Positive
Negative
Positive
Negative
Condition
Test
True positive
False negative
False positive
True negative

# DANN4

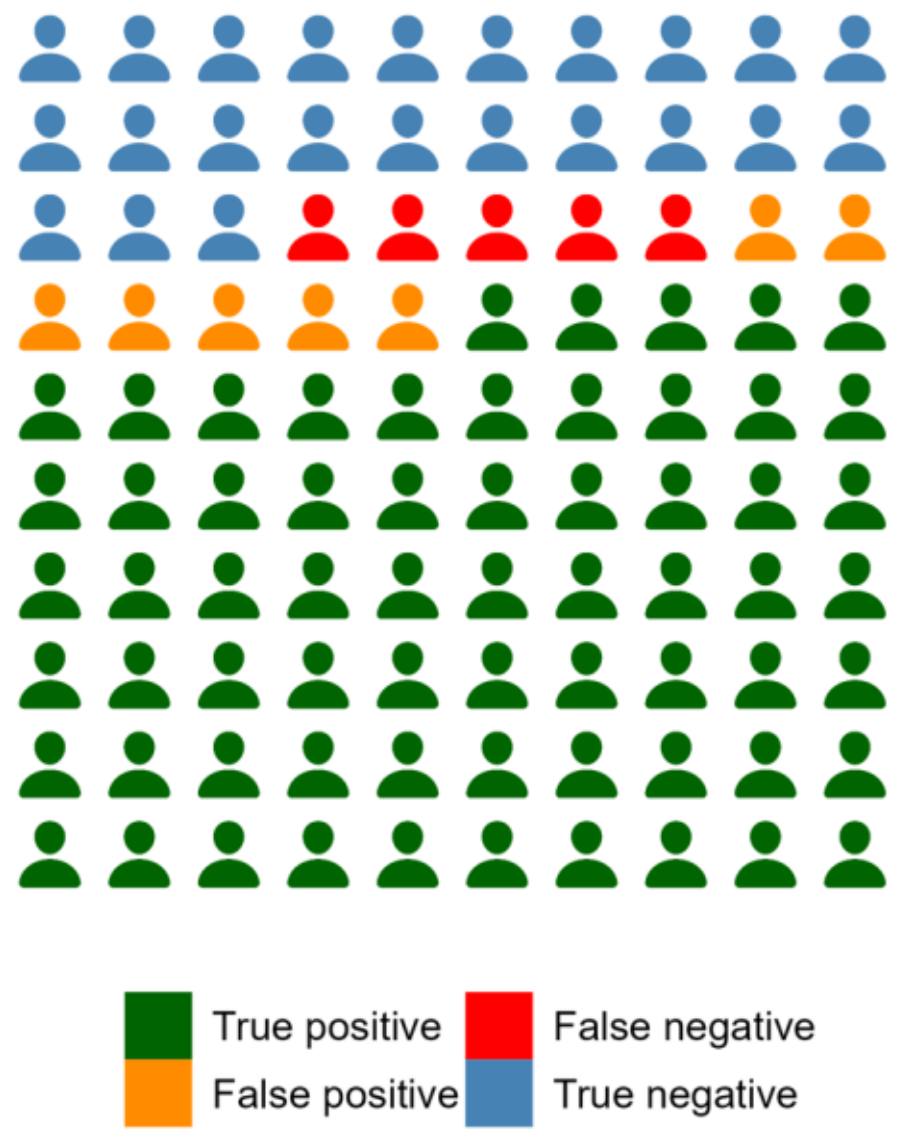
True positive
False negative
False positive
True negative

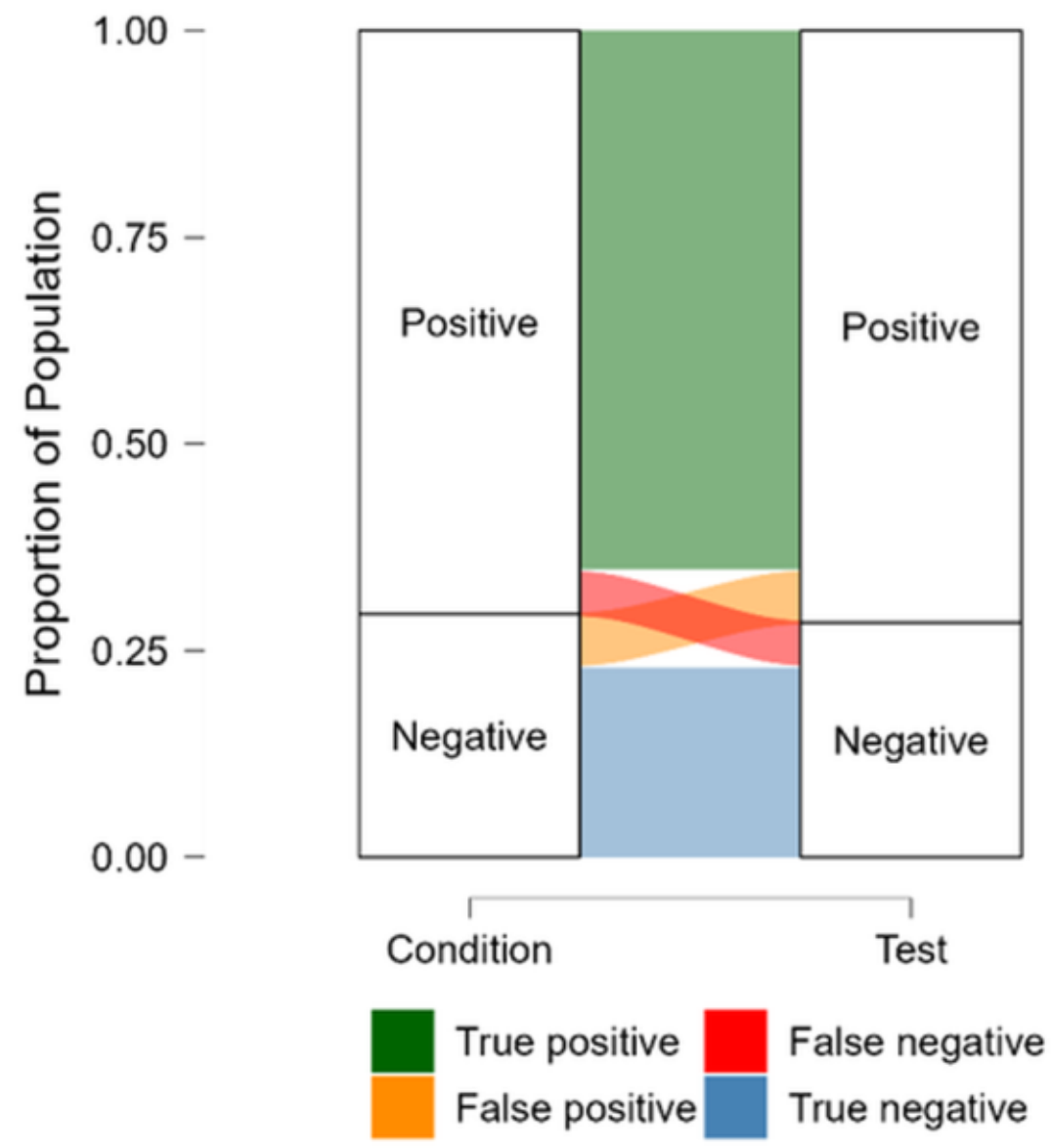
1.00
0.75
0.50
0.25
0.00
Proportion of Population
Positive
Positive
Negative
Negative
Condition
Test
True positive
False negative
False positive
True negative

# DANN5

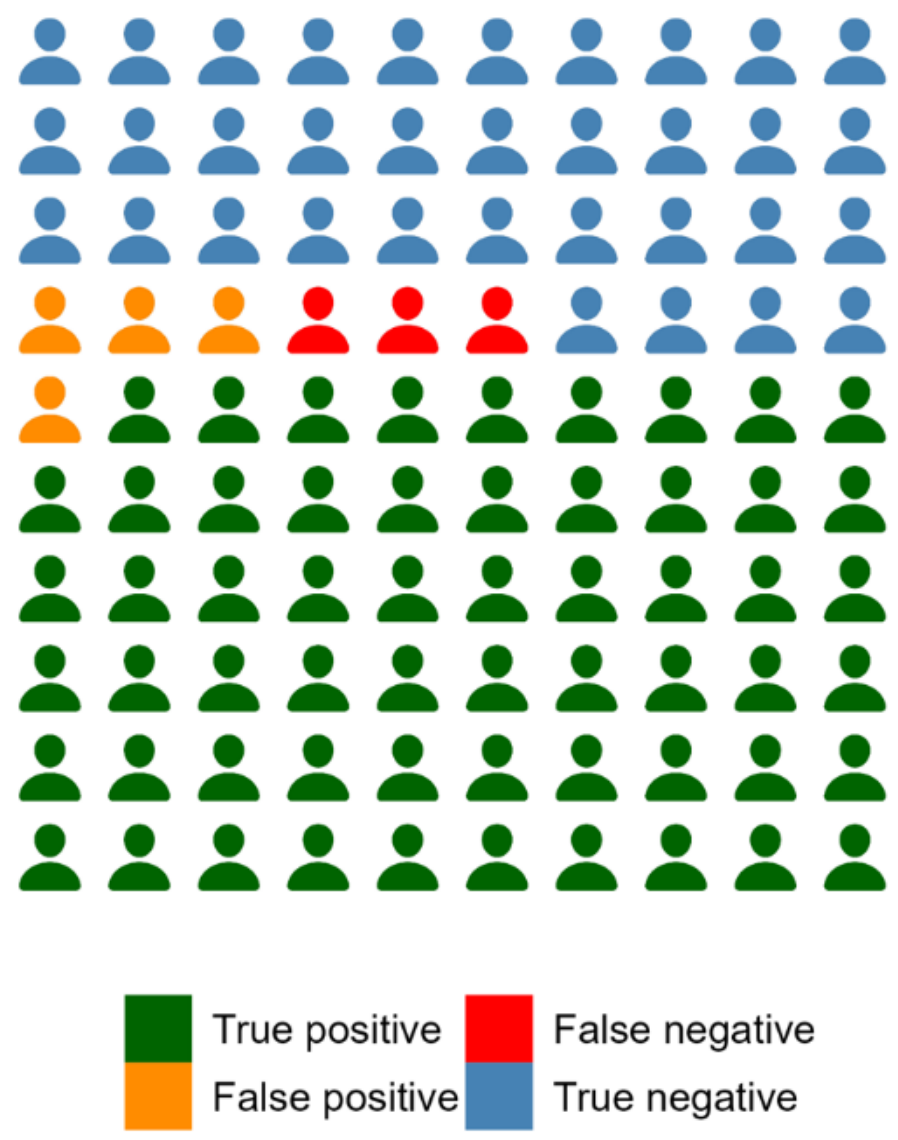
True positive
False negative
False positive
True negative

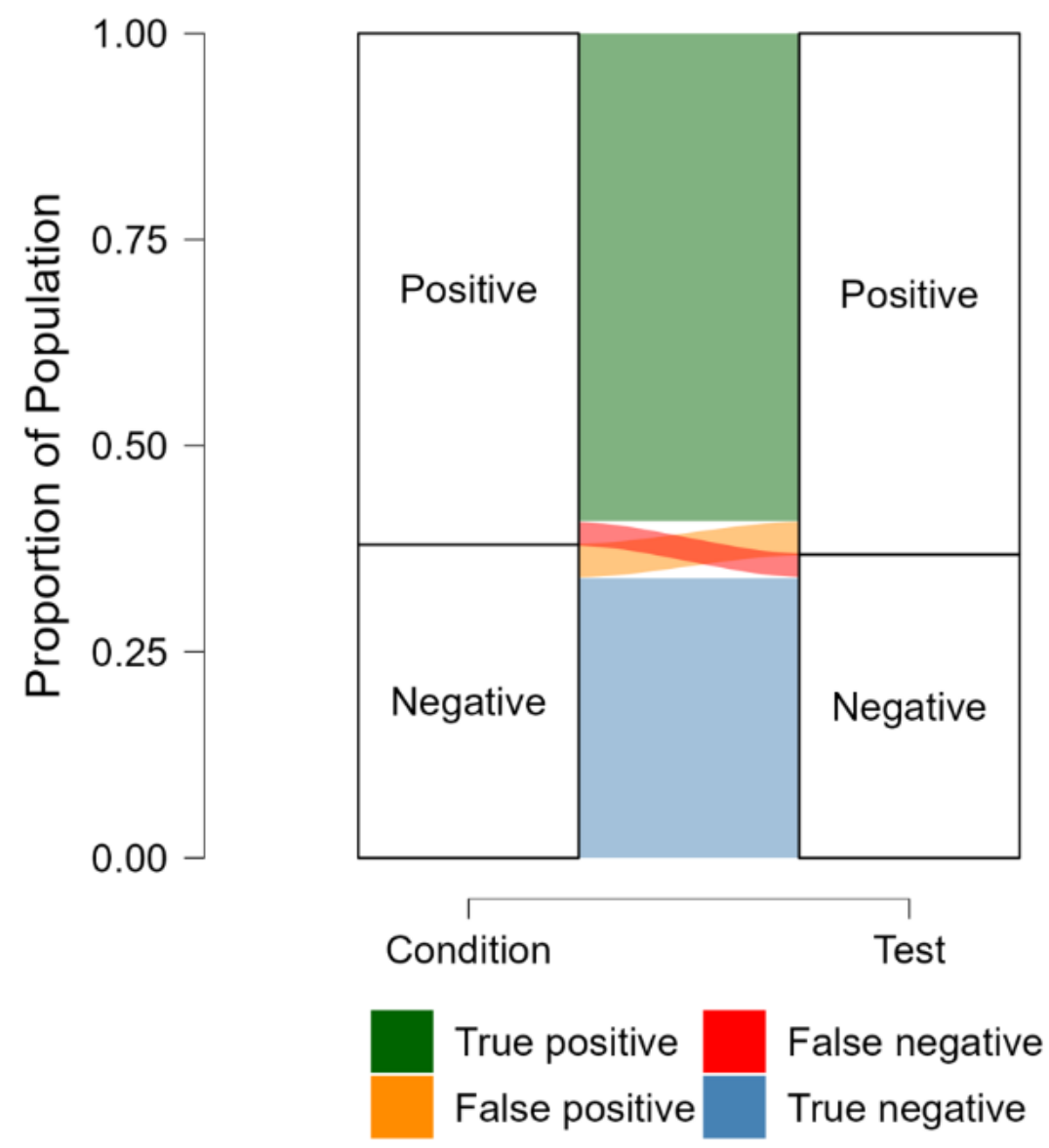
1.00
0.75
0.50
0.25
0.00
Proportion of Population
Positive
Positive
Negative
Negative
Condition
Test
True positive
False negative
False positive
True negative

## DANN6

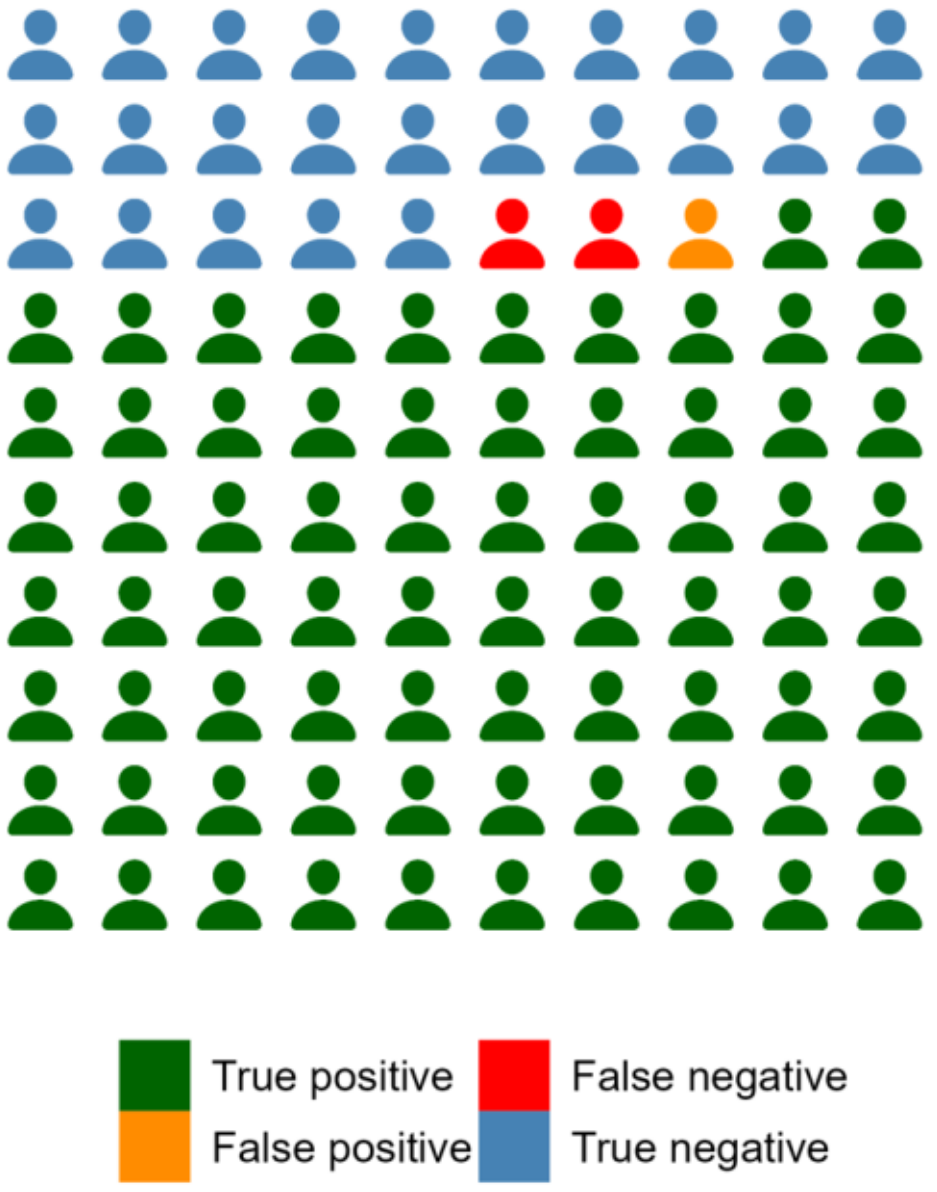
True positive
False negative
False positive
True negative

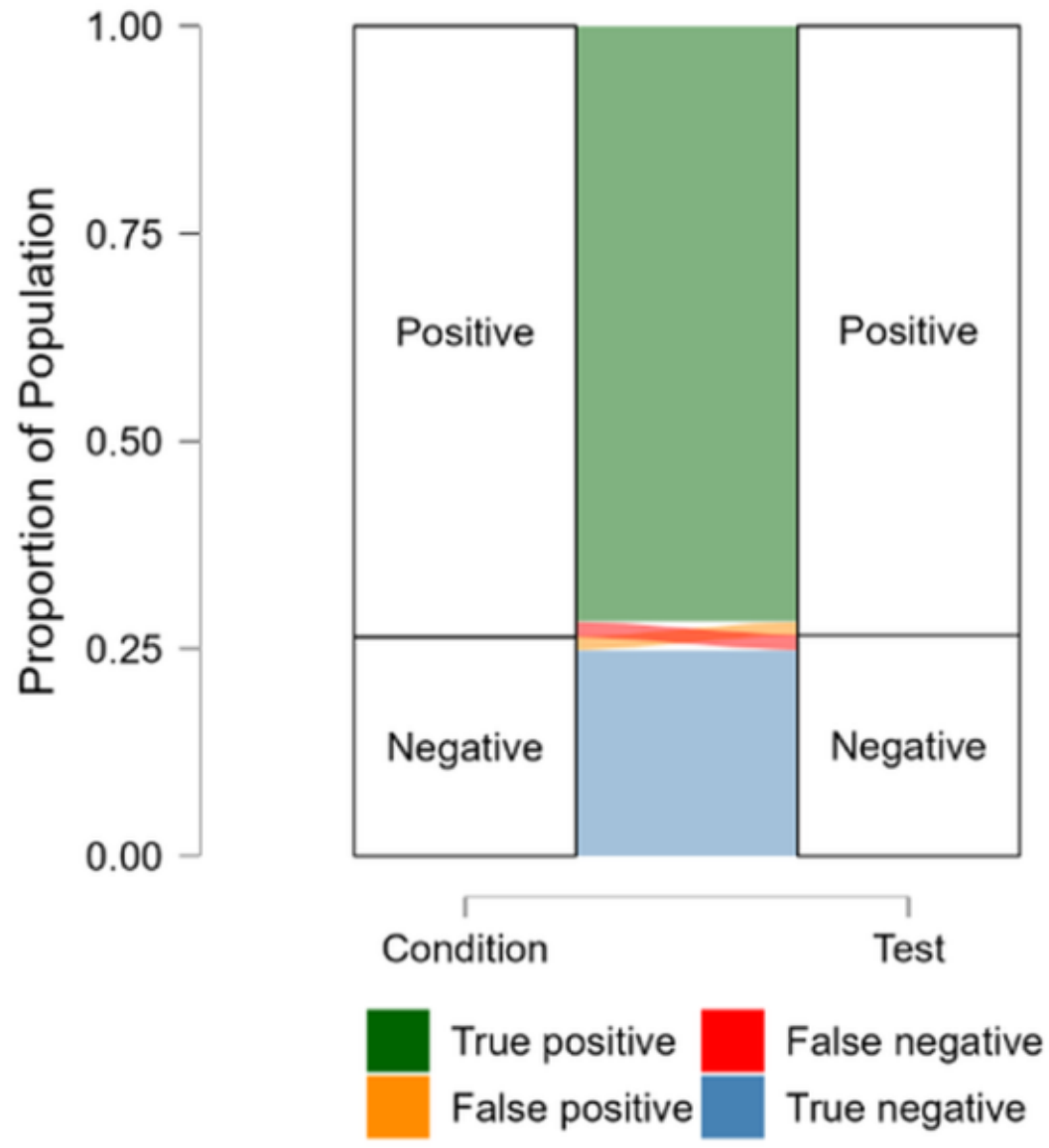
1.00
0.75
0.50
0.25
0.00
Proportion of Population
Positive
Negative
Positive
Negative
Condition
Test
True positive
False negative
False positive
True negative

**Supplementary Table 3 Pairwise Comparison of AUC Statistics Across DANN Models in External Validation**

| Pairwise comparison | AUC | p value | z value | 95% CI |
|---|---|---|---|---|
| **DANN1 vs DANN2** | 0.966 / 0.976 | 0.5239 | 0.637 | -0.0211 to 0.0414 |
| **DANN1 vs DANN3** | 0.966 / 0.972 | 0.6757 | 0.418 | -0.0219 to 0.0337 |
| **DANN1 vs DANN4** | 0.966 / 0.954 | 0.5371 | 0.617 | -0.0273 to 0.0524 |
| **DANN1 vs DANN5** | 0.966 / 0.871 | 0.0012 | 3.233 | 0.0376 to 0.153 |
| **DANN1 vs DANN6** | 0.966 / 0.793 | <0.0001 | 6.141 | 0.118 to 0.229 |
| **DANN2 vs DANN3** | 0.976 / 0.972 | 0.7827 | 0.276 | -0.0258 to 0.0343 |
| **DANN2 vs DANN4** | 0.976 / 0.954 | 0.2824 | 1.075 | -0.0187 to 0.0641 |
| **DANN2 vs DANN5** | 0.976 / 0.871 | <0.0001 | 4.187 | 0.0562 to 0.155 |
| **DANN2 vs DANN6** | 0.976 / 0.793 | <0.0001 | 5.651 | 0.120 to 0.247 |
| **DANN3 vs DANN4** | 0.972 / 0.954 | 0.3806 | 0.877 | -0.0228 to 0.0598 |
| **DANN3 vs DANN5** | 0.972 / 0.871 | 0.0003 | 3.599 | 0.0462 to 0.157 |
| **DANN3 vs DANN6** | 0.972 / 0.793 | <0.0001 | 5.709 | 0.118 to 0.241 |
| **DANN4 vs DANN5** | 0.954 / 0.871 | 0.0125 | 2.497 | 0.0178 to 0.148 |
| **DANN4 vs DANN6** | 0.954 / 0.793 | <0.0001 | 5.351 | 0.102 to 0.220 |
| **DANN5 vs DANN6** | 0.871 / 0.793 | 0.0436 | 2.018 | 0.00225 to 0.154 |

**Supplementary Figure 3 ROC Curves for Senior and Junior Graders**

The senior grader had an AUC of 0.999 (SE = 0.000365; 95% CI: 0.988–1.000), and the junior grader had an AUC of 0.933 (SE = 0.0146; 95% CI: 0.901–0.957); the difference between AUCs was 0.0665 (SE = 0.0145; 95% CI: 0.0381–0.0949; z = 4.587; p < 0.0001).

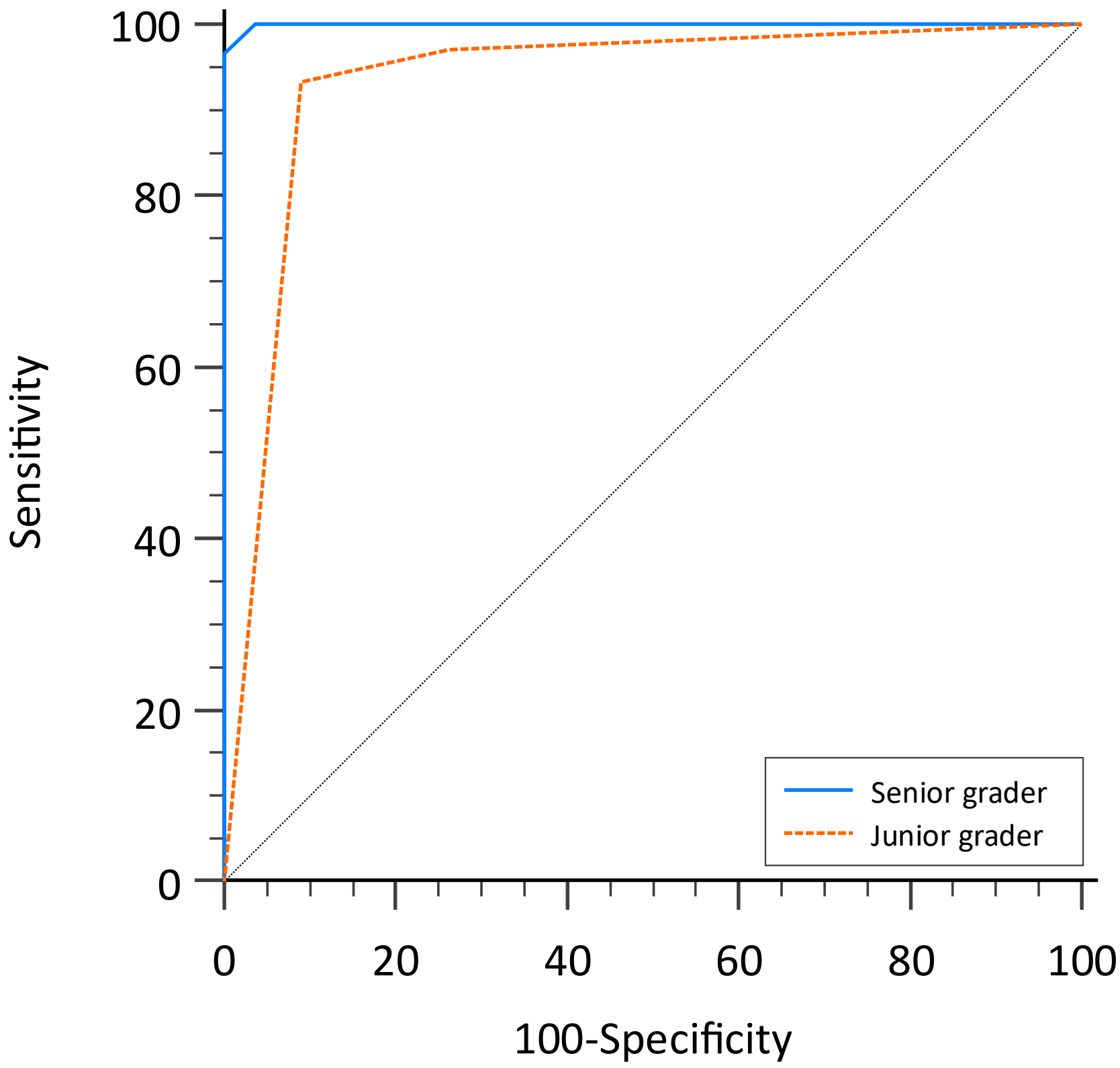